\pdfoutput=1

\documentclass[11pt]{article}

\usepackage[final]{acl2025}

\usepackage{times}
\usepackage{latexsym}
\usepackage{adjustbox}
\usepackage{listings}

\usepackage[T1]{fontenc}

\usepackage[utf8]{inputenc}

\usepackage{microtype}

\usepackage{inconsolata}

\usepackage{graphicx}

\usepackage{graphicx}
\usepackage{multirow}
\usepackage{makecell}
\usepackage{color,soul}
\usepackage{float}
\usepackage{xcolor,colortbl}
\restylefloat{table}
\usepackage{enumitem}
\usepackage{multirow}
\usepackage{multicol}
\usepackage{array}
\usepackage{soul}
\usepackage{hyperref}
\usepackage{amsthm}
\usepackage{amsmath}
\theoremstyle{definition}

\usepackage{framed}
\usepackage{algorithm}
\usepackage{algpseudocode}

\usepackage{amssymb}
\usepackage{enumitem}
\usepackage{bbm}

\usepackage{arydshln}

\usepackage{siunitx}

\usepackage{tabularx}
\usepackage{tipa} 

\usepackage{pifont}   

\usepackage{xspace}

\usepackage{tikz}

\usepackage{calc} 

\usepackage{listings}
\usepackage{subcaption}

\newcommand{\cudr}{\textsc{CuDR}\xspace}

\definecolor{darkyellow}{rgb}{0.95,0.75,0.05}  
\definecolor{darkgreen}{rgb}{0,0.39,0}         
\definecolor{deeppurple}{rgb}{0.7,0,0.85}      

\definecolor{darkgreen}{RGB}{0,100,0}

\definecolor{lightgray}{RGB}{169,169,169}

\newcommand{\LThree}{\textcolor{red}{L3}\xspace}
\newcommand{\LTwo}{\textcolor[rgb]{0.95,0.75,0.05}{L2}\xspace} 
\newcommand{\LOne}{\textcolor[rgb]{0,0.39,0}{L1}\xspace}    
\newcommand{\LZero}{\textcolor{blue}{L0}\xspace}
\newcommand{\LOwn}{\textcolor[rgb]{0.7,0,0.85}{Own}\xspace}  
\newcommand{\IOI}{\textcolor{gray}{IOI}\xspace}
\newcommand{\random}{\textcolor{lightgray}{random}\xspace}
\newcommand{\Random}{\textcolor{lightgray}{Random}\xspace}

\title{Discursive Circuits: How Do Language Models Understand Discourse Relations?}

\author{Yisong Miao \quad \quad Min-Yen Kan \\
Web IR / NLP Group (WING), National University of Singapore \\ 
\texttt{\{yisong, kanmy\}@comp.nus.edu.sg}
}

\begin{document}
\maketitle

\begin{abstract}

Which components in transformer language models are responsible for discourse understanding? We hypothesize that sparse computational graphs, termed as \textit{discursive circuits}, control how models process discourse relations. Unlike simpler tasks, discourse relations involve longer spans and complex reasoning. To make circuit discovery feasible, we introduce a task called Completion under Discourse Relation (\cudr), where a model completes a discourse given a specified relation. To support this task, we construct a corpus of minimal contrastive pairs tailored for activation patching in circuit discovery. Experiments show that sparse circuits ($\approx 0.2\%$  of a full GPT-2 model) recover discourse understanding  in the English PDTB-based \cudr task. 

These circuits generalize well to unseen discourse frameworks such as RST and SDRT. 
Further analysis shows lower layers capture linguistic features such as lexical semantics and coreference, while upper layers encode discourse-level abstractions. Feature utility is consistent across frameworks (e.g., coreference supports Expansion-like relations).

\end{abstract}

\section{Introduction}

\begin{figure}[t!]
\includegraphics[width=7.6cm]{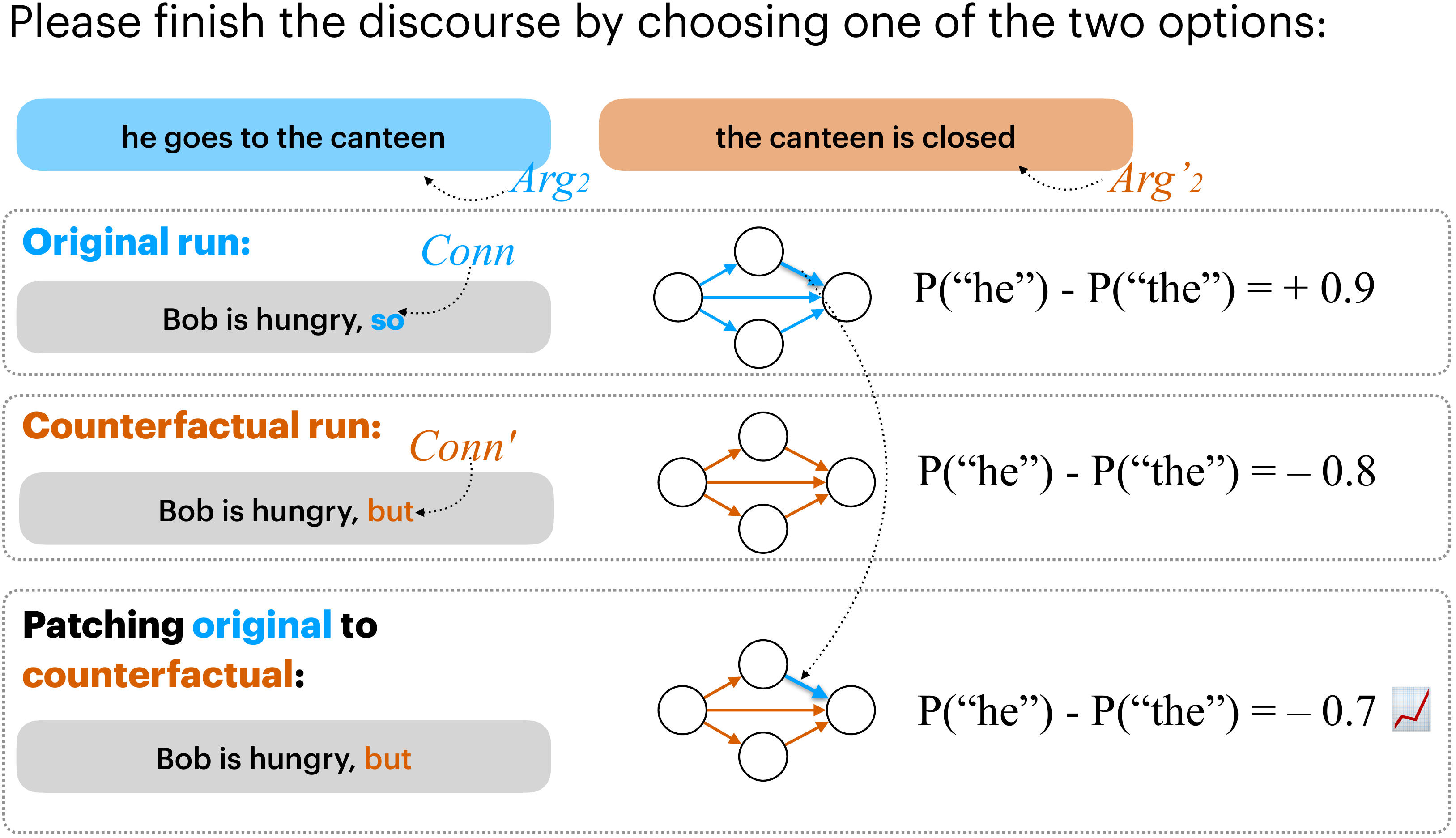}
\centering
\caption{\textbf{Task Overview:} The \cudr task enables discovery of discursive circuits by contrasting model predictions under minimal changes to the discourse connectives. Activation patching reveals components causally responsible for shifting the model’s prediction.}
\label{fig: intro}
\end{figure}

Discourse structure is essential for ensuring language models (LMs) to behave safely and ethically \cite{kim2025alignstructurealigninglarge, nakshatri2025talkingpointbasedideological}. Yet, little is known about how discourse is internally processed by LMs, limiting our ability to guarantee that they are reliable and free from harmful outputs.
Transformer circuit discovery \cite{zhang2024towards} is a promising method that identifies sparse computational subgraphs causally responsible for specific behaviors. Unlike attention visualization \cite{jain-wallace-2019-attention} or rationale generation \cite{wiegreffe2021teach}, circuits provide mechanistic, intervention-based explanations that reveal which components causally drive the model’s output.
Existing circuit discovery methods focus on simple tasks, like numeric comparison \cite{hanna2023how} which is well-suited for next-word prediction (e.g. ``The year after 1731 is $\rightarrow$''). 
In contrast, discourse relations involve longer contexts and more complex reasoning, making direct adaptation of existing methods infeasible.

We contribute a key insight: by bridging the linguistic structure of discourse and the requirements of circuit discovery, we open a new path for mechanistic understanding of complex language tasks.
On the discourse side, we hold the initial argument $Arg_1$ (e.g. ``Bob is hungry'', Figure~\ref{fig: intro}) unchanged and introduce a counterfactual connective $Conn'$ (e.g., \textcolor{orange}{``but''}) that prompts the model to select an alternative continuation \textcolor{orange}{$Arg_2'$} (``the canteen is closed''), which is only coherent under the counterfactual discourse relation.
On the circuit discovery side, the method relies on minimal contrastive pairs, where inputs differ slightly but yield significantly different outputs.
To identify influential model components, we patch activations \cite{nanda2023attribution} from the original run into the counterfactual run and observe changes in prediction. The resulting discursive circuits are composed of connections with significant causal influence.

To support this task, we construct a dataset spanning major discourse frameworks, including Penn Discourse Treebank (PDTB; \citeauthor{webber2019penn},\citeyear{webber2019penn}), Rhetorical Structure Theory (RST; \citeauthor{mann1987rhetorical},\citeyear{mann1987rhetorical}), and Segmented Discourse Representation Theory (SDRT; \citeauthor{asher2003logics},\citeyear{asher2003logics}). Each instance contains an original annotation from the source corpus, along with a set of counterfactual connectives and their alternative completions. The three frameworks have 10 to 17 distinct discourse relations each, and together contribute a total of 27,754 instances.

Using our datasets, we discover discursive circuits in the GPT-2 medium model.
For most discourse relations, the identified circuits achieve around 90\% faithfulness while involving only 0.2\% of model connections.
We show that circuits derived from PDTB generalize well to unseen discourse frameworks such as RST and SDRT, suggesting that language models may encode a shared representation of discourse relations. 
We also construct a novel circuit hierarchy adapted from PDTB’s three-level taxonomy. To our knowledge, this is the first discourse hierarchy grounded in neural circuit components. 
Together, our circuits and hierarchy provide a new form of discourse representation, enabling direct cross-framework comparison and fine-grained decomposition into linguistic features. We discover similar utilities across different frameworks (e.g., coreference is prominent in all Expansion-like relations) \footnote{The software and data are publicly available at: \url{https://github.com/YisongMiao/Discursive-Circuits}.}. 

\section{Circuit Discovery with \cudr}
\label{sec: CuDR}
We propose a generic workflow to dissect a language model's discourse understanding via circuit discovery, which is compatible with any discourse framework.
We introduce the Completion under Discourse Relation task (\cudr, pronounced ``koo-der''), where $Arg_1$ remains fixed, while the connective is swapped (\textcolor{blue}{$Conn$} $\rightarrow$ \textcolor{orange}{$Conn'$}), requiring the model to shift its prediction from \textcolor{blue}{$Arg_2$} to \textcolor{orange}{$Arg_2'$}. 

\subsection{Completion under Discourse Relation}
\begin{table}[th!]
\small
\begin{tabular}{|p{7.2cm}|}
\hline
\textbf{Input:} \\
$d_{ori} = (Arg_{1}, \textcolor{blue}{Arg_{2}}, \textcolor{blue}{R}, \textcolor{blue}{Conn})$ \\
$d_{cf} = (Arg_{1}, \textcolor{orange}{Arg_{2}'}, \textcolor{orange}{R'}, \textcolor{orange}{Conn'})$
\\ \hline \hline
\textbf{\cudr Task (Original):} \\
\texttt{Please finish the discourse by choosing one of the two options:}
$\textcolor{blue}{Arg_{2}}$ or $\textcolor{orange}{Arg_{2}'}$ \\
\texttt{To complete:} $Arg_{1}, \textcolor{blue}{Conn}$ \\ \hline
\textbf{Correct answer:} $\textcolor{blue}{Arg_{2}}$, \textbf{Incorrect answer: }$\textcolor{orange}{Arg_{2}'}$ \\ \hline
\textbf{Example:} \texttt{Please finish the discourse by choosing one of the two options:} \textcolor{blue}{``he goes to the canteen''} or \textcolor{orange}{``the canteen is closed''} \\
\texttt{To complete:}
[Bob is hungry]$_{Arg_{1}}$ \textcolor{blue}{[so]$_{Conn}$ $\Rightarrow$ [he goes to the canteen]$_{Arg_{2}}$}
\\ \hline \hline
\textbf{\cudr Task (Counterfactual):} \\
\texttt{Please finish the discourse by choosing one of the two options:}
$\textcolor{blue}{Arg_{2}}$ or $\textcolor{orange}{Arg_{2}'}$  \\
\texttt{To complete:} $Arg_{1}, \textcolor{orange}{Conn'}$ \\ \hline
\textbf{Correct answer:} $\textcolor{orange}{Arg_{2}'}$, \textbf{Incorrect answer:} $\textcolor{blue}{Arg_{2}}$ \\ \hline
\textbf{Example:} \texttt{Please finish the discourse by choosing one of the two options:} \textcolor{blue}{``he goes to the canteen''} or \textcolor{orange}{``the canteen is closed''} \\
\texttt{To complete:}
[Bob is hungry]$_{Arg_{1}}$ \textcolor{orange}{[but]$_{Conn'} \Rightarrow$ [the canteen is closed]$_{Arg_{2}'}$}
\\ \hline
\end{tabular}
\centering
\caption{\textbf{Formalization of the \cudr task:} the model must complete the discourse by either $\textcolor{blue}{Arg_{2}}$ or the counterfactual $\textcolor{orange}{Arg_{2}'}$, based on which best fits as a continuation of $Arg_{1}$ following \textcolor{blue}{$Conn$} or \textcolor{orange}{$Conn'$} (best in color).}
\label{tab: formalization}
\end{table}
\cudr creates a controlled environment to test a model's discursive behavior. By simply altering the discourse connective (from \textcolor{blue}{original (ori)} to \textcolor{orange}{counterfactual (CF)}; Table~\ref{tab: formalization}), the model's continuation shifts sharply in response. For example, in the original discourse, a Contingency relation is expressed with the connective \textcolor{blue}{``so''}, leading to a completion that \textcolor{blue}{``he goes to the canteen''}. However, when the discourse relation is shifted to a counterfactual Comparison relation (signaled by \textcolor{orange}{``but''}), the model should sharply change its prediction to an argument that negates the expectation of eating (i.e., \textcolor{orange}{``the canteen is closed''}). Note that while circuit discovery has been applied under various settings \cite{zhang2024towards}, we adopt such a setup to steer the model, because it captures the dynamic nature of discourse understanding. 

Concretely, the original discourse consists of two arguments, $Arg_1$ and $Arg_2$, linked by a discourse relation $R$ and connective $Conn$, formally denoted as $d_{\text{ori}} = (Arg_1, Arg_2, R, Conn)$. The counterfactual instance, $d_{\text{cf}} = (Arg_1, Arg_2', R', Conn')$, preserves $Arg_1$ but substitutes the continuation and relation ($R' \neq R$), forming a minimal contrastive pair required by activation patching.

\subsection{Circuit Discovery}

\paragraph{Activation Patching.}  Transformer circuits are computational graphs that model the information flow from an input token, through residual flow among intermediate nodes (i.e., MLP layers and attention heads) to the output probability of the next token. To identify influential connections inside the circuits, we intervene in the model by replacing the activation of a counterfactual (corrupted) run by the activation of an original (clean) run. 
\begin{equation}
    g(e) = L(x_{cf} | do(E=e_{ori})) - L(x_{cf})
    \label{eq: activation-patching}
\end{equation}

Concretely, we define the impact of introducing an intervening edge $e$ (denoted by $g(e)$) as the difference in a metric $L$ when patching the activation of edge $e$ from the original run ($do(E=e_{ori})$). Formally, $g(e)$ is computed as the difference between $L(x_{cf} | do(E=e_{ori}))$ where $e$ is restored to its clean value, and $L(x_{cf})$, the metric value under the corrupted run. 

\paragraph{Accelerate by Attribution Patching.} To overcome the low speed and inference costs for activation patching \cite{conmy2023towards}, we adopt a first order Taylor approximation to Equation \ref{eq: activation-patching} and use the Edge Attribution Patching (EAP) method \cite{nanda2023attribution, syed-etal-2024-attribution}. For an edge $e=(u, v)$, the change of metric $g(e)$ is:

\begin{equation}
    g(e) \approx (z_u^{ori} - z_u^{cf})^\top \nabla_{v} L(x_{cf}),
\end{equation}

\noindent where $z_u^{ori}$ and $z_u^{cf}$ denote the activation at node $u$ in the original or counterfactual runs, and $\nabla_{v} L(x_{cf})$ is the gradient of metric $L$ at node $v$. With the approximation, we can now calculate $g(e)$ for all edges by two forward passes and one backward pass, greatly enhancing efficiency (by a factor of $10^3$ in our practice), while preserving the performance of circuits \cite{syed-etal-2024-attribution}.

\begin{figure}[h]
\includegraphics[width=7.6cm]{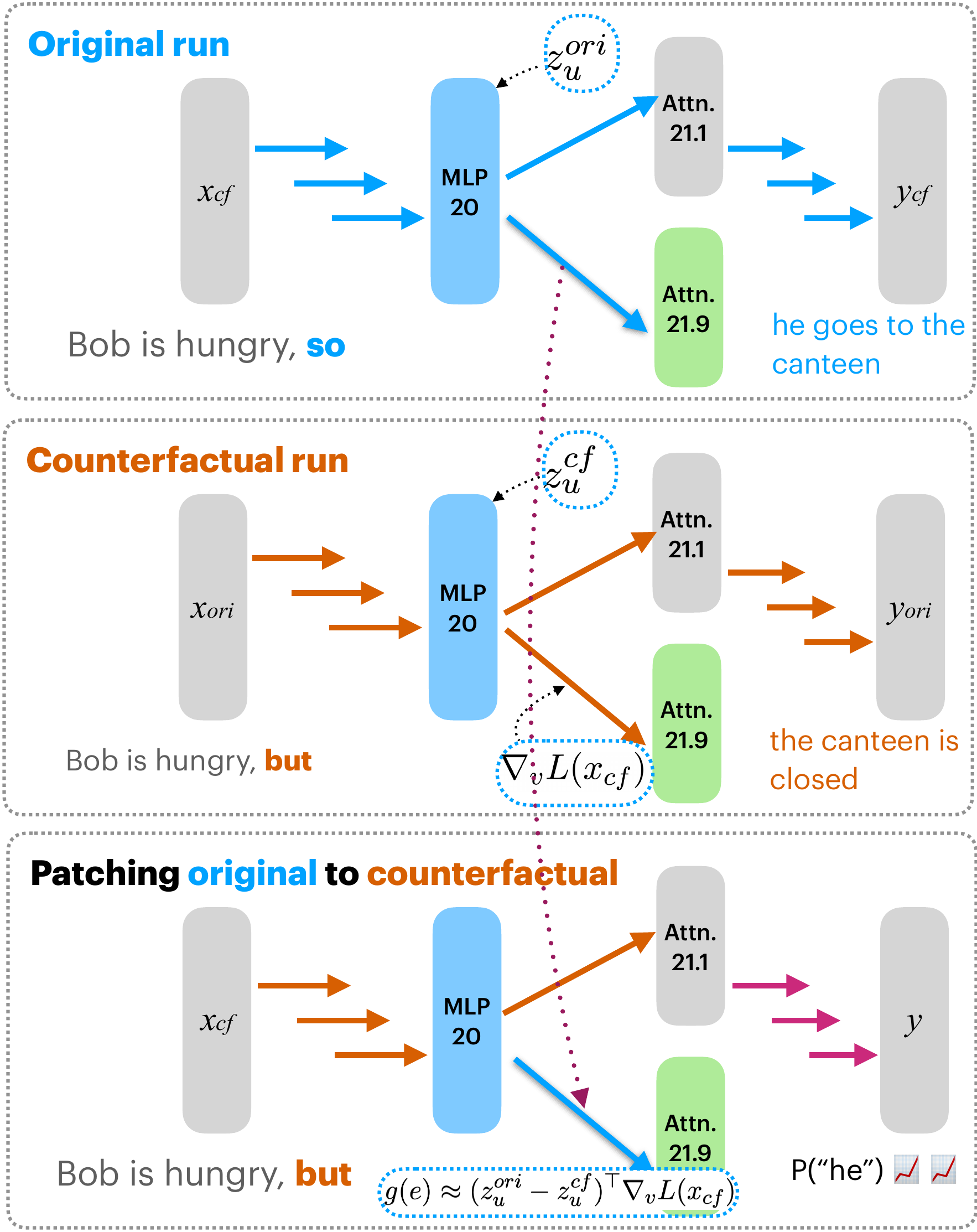}
\centering
\caption{\textbf{Illustration of attribution patching with \cudr:} We steer the model's prediction from the \textcolor{orange}{counterfactual} toward the \textcolor{blue}{original} outcome. 
Activations from the \textcolor{blue}{original run} are patched into the \textcolor{orange}{counterfactual run} to influence the model’s prediction. 
}
\label{fig: patching}
\end{figure}

\paragraph{Attribution Patching Using \cudr.}

We first input the model with the \textcolor{orange}{counterfactual (CF)} input, and the model produces a CF output. Using the same CF input, we then perform activation patching from the \textcolor{blue}{original (Ori)} to restore the model's prediction to the Ori output.
In the CF run, the model receives $\textcolor{orange}{x_{cf}}$, constructed from $Arg_{1}$ and a counterfactual discourse connective ($\textcolor{orange}{Conn'}$). The correct prediction is the counterfactual completion ($\textcolor{orange}{Arg_{2}'}$). In the \textcolor{blue}{ori} run, the model receives $\textcolor{blue}{x_{ori}}$ as input, which consists of ($Arg_{1}, \textcolor{blue}{Conn}$). The correct output is the original \textcolor{blue}{$Arg_{2}$}.
Attribution patching (Figure~\ref{fig: patching}) works by replacing activations from the \textcolor{orange}{CF} run with those from the \textcolor{blue}{Ori} run. For example, to measure the importance of the edge between MLP 20 and Attention Head 21.9 (Attn. 21.9), we replace the activation flowing from MLP 20 into Attn. 21.9 with the corresponding activation from the \textcolor{blue}{Ori} run and observe $g(e)$, which is the change in the model's output. 

\paragraph{Construct Discursive Circuits.} The discursive circuit for a given discourse relation is constructed by applying attribution patching to the \cudr task over a set of samples for that relation. We compute the average $g(e)$ for each edge and select those with the highest absolute $g(e)$ values as the most important. In practice, the top 1000 such edges are sufficient to steer the model faithfully, similar to prior work \cite{hanna2024have}.

\subsection{The \cudr Dataset}
\label{sec: cudr datasets}

We construct an augmented dataset by prompting a large language model (LLM) with the original $Arg_{1}$ and a counterfactual \textcolor{orange}{$Conn'$}, along with detailed instructions and discourse relation definitions (Appendix~\ref{app: dataset construction}). We employ GPT-4o-mini for its good instruction-following ability and lower cost.

\begin{table}[th]
\footnotesize
\centering
\resizebox{\columnwidth}{!}{%
\begin{tabular}{lll}
\textbf{Discourse Relation} & \textbf{Ori Connective} & \textbf{CF Connective} \\ 
\hline
\textbf{Comparison.Concession.Arg2-as-denier} & however & \makecell[l]{because \\ for example} \\
\hline
\textbf{Comparison.Contrast} & by comparison & \makecell[l]{specifically \\ in other words} \\
\hline
\textbf{Contingency.Reason} & because & \makecell[l]{so \\ however} \\
\hline
\textbf{Contingency.Result} & so & \makecell[l]{because \\ by comparison} \\
\hline
\textbf{Expansion.Conjunction} & and & \makecell[l]{however \\ so} \\
\hline
\textbf{Expansion.Equivalence} & in other words & \makecell[l]{however \\ for example} \\
\hline
\textbf{Expansion.Instantiation.Arg2-as-instance} & for example & \makecell[l]{because \\ however} \\
\hline
\textbf{Expansion.Level-of-detail.Arg1-as-detail} & in short & \makecell[l]{however \\ so} \\
\hline
\textbf{Expansion.Level-of-detail.Arg2-as-detail} & specifically & \makecell[l]{instead \\ by comparison} \\
\hline
\textbf{Expansion.Substitution.Arg2-as-subst} & instead & \makecell[l]{because \\ in other words} \\
\hline
\textbf{Temporal.Asynchronous.Precedence} & then & \makecell[l]{however \\ previously} \\
\hline
\textbf{Temporal.Asynchronous.Succession} & previously & \makecell[l]{so \\ then} \\
\hline
\textbf{Temporal.Synchronous} & while & \makecell[l]{so \\ then} \\
\hline
\end{tabular}%
}
\caption{\textbf{\cudr Dataset:} PDTB's discourse relations with corresponding original (Ori) connectives and counterfactual (CF) connectives (subset displayed for CF).}
\label{tab:conn}
\end{table}

Building on the taxonomy of counterfactual discourse relations proposed by \citet{miao-etal-2024-discursive}, our \cudr dataset adopts a PDTB3-based design (Table~\ref{tab:conn}). For each discourse relation alongside its \textcolor{blue}{original} connective, we construct five \textcolor{orange}{counterfactual} discourse connectives. For example, the Comparison.Concession.Arg2-as-denier relation (e.g., \textcolor{blue}{``however''}, Row~1 in Table~\ref{tab:conn}) is considered counterfactual to both a Contingency relation (signaled by \textcolor{orange}{``because''}) and an Instantiation relation (\textcolor{orange}{``for example''}). We provide a complete list of connectives and their mappings in Appendix~\ref{app: cf conn}.

\begin{table}[th]
\centering
\small
\resizebox{\columnwidth}{!}{%
\begin{tabular}{lrr}
\textbf{Discourse framework} & \textbf{\# of DR} & \textbf{\# of CuDR data} \\ \hline
\textbf{PDTB} & 13 & 11,843 \\
\textbf{GDTB} & 12 & 5,253 \\
\textbf{GUM-RST} & 17 & 6,805 \\
\textbf{SDRT} & 10 & 3,853 \\ \hline
\textbf{Total} &  & 27,754 \\ \hline
\end{tabular}%
}
\caption{\textbf{CuDR Dataset Statistics:} Number of unique discourse relations and CuDR data across frameworks.}
\label{tab:cudr-datasets-by-framework}
\end{table}

We extend our dataset construction beyond PDTB to include additional corpora: the GUM Discourse Treebank (GDTB; \citealt{liu-etal-2024-gdtb}), a more up-to-date PDTB-style corpus, as well as GUM-RST \cite{zeldes2017gum} and SDRT \cite{asher2003logics}. To enable the generation of counterfactual instances from non-PDTB corpora, we construct relation mappings from RST to PDTB (Table \ref{tab:rst2pdtb}) and from SDRT to PDTB (Table \ref{tab:sdrt2pdtb} in Appendix \ref{app: dataset}). For example, SDRT's \textcolor{blue}{Explanation} relation is mapped to PDTB’s \textcolor{blue}{Contingency.Cause.Reason}, then its corresponding counterfactual relations \textcolor{orange}{Result (``so'')} and \textcolor{orange}{Contrast (``however'')} are found in the PDTB-based taxonomy.

Table~\ref{tab:cudr-datasets-by-framework} summarizes the metadata per discourse framework. Each original and counterfactual discourse pair, $(d_{\text{ori}}, d_{\text{cf}})$, is treated as a single data instance in the \cudr dataset. For each discourse relation in each corpus, we sample up to 50 original instances for circuit discovery and evaluation. With five counterfactual connectives per relation, this yields up to 250 \cudr instances. We discard minority relations with fewer than 20 instances, as well as low-quality instances where $Arg_2$ and $Arg_2'$ are overly similar.
We consider our sample size sufficient, as \citet{yao2024knowledge} use a median of only 52. 
To validate the automated constructions, one author manually verified 40 \cudr samples and found them all valid as an indicative evaluation, with \textcolor{orange}{$Arg'_2$} coherent with \textcolor{blue}{$Arg_1$} and \textcolor{orange}{$Conn'$}. The language in \textcolor{orange}{$Arg'_2$} tends to be straightforward, but it is desired because we want salient relations. 
We also construct a small set of counterfactual \textcolor{orange}{$Arg'_2$} instances, written by the first author, for indicative comparison (Appendix \ref{app: human annotated cf}).
Preliminary trials with open-source Llama-3.1-8B-Instruct \cite{grattafiori2024llama} to generate \cudr data were unsuccessful as the model did not follow our task instruction. 

\section{Evaluate Discursive Circuits}

\definecolor{darkgreen}{rgb}{0.0,0.5,0.0}
\definecolor{MyRed}{RGB}{220,20,60}    
\definecolor{MyPurple}{RGB}{128,0,128} 

We conduct our evaluation to answer the following research questions (RQs): 

\noindent\textbf{RQ1:} Do discursive circuits faithfully recover the full model's performance? \\
\textbf{RQ2:} Do discursive circuits generalize across different discourse frameworks and relation types?  \\
\textbf{RQ3:} Are discursive circuits composed of components associated with specific linguistic features?

\paragraph{Implementation Details.}
Following \citet{hanna2024have, mondorf2025circuitcomposition}, we focus on a single model for in-depth analysis and adopt their choice of GPT-2 medium \cite{radford2019language} for its manageable memory requirements. 
To identify circuits for specific discourse relations, we use a sample size of 32 for both circuit discovery and validation, and apply the standard practice of using the batch mean for node value patching \cite{miller2024transformer}.
We repeat each experiment five times with different data samples and average the outcomes
for stability. Before circuit discovery, we fine-tune the model on held-out \cudr data (half of the PDTB subset) to align it with our task setting and ensure it follows the intended instructions (Appendix~\ref{app: cudr model finetune}). Our fine-tuned model is not perfect, achieving around $80\%$ accuracy in our \cudr task. However, we use the entire dataset (including incorrectly predicted instances) for both circuit discovery and evaluation to fully capture the distribution of the task. 
Aside from GPT-2 medium, we also scale our experiments to GPT-2 large and find that the larger model has similar performance (Appendix~\ref{app: scaling}). 

\noindent\textbf{Baseline Circuits:} 
(1) Following \citet{hsu2025efficient, basu2025on}, we benchmark \textbf{random circuits} on our \cudr task, where circuit edges are sampled randomly from the transformer without any learned importance. This comparison evaluates whether our learned circuits provide advantages beyond random selection. 
(2) We also replicate the \textbf{Indirect Object Identification (IOI)} circuit \cite{wang2023interpretability} in our own model as a baseline circuit. In the IOI task, the model is given a prompt like ``John and Mary went to a bar. Mary gave a beer to'', and should predict ``John''. This circuit represents the model's general next-word prediction ability, without discourse-specific reasoning. Comparing against IOI allows us to test whether discursive circuits capture discourse-specific computation beyond standard language modeling.

\paragraph{Evaluation Metric.}
Our metric follows \citet{miller2024transformer} to calculate the \textbf{logit difference} between the correct and incorrect answers. Specifically, we treat the original discourse's \textcolor{blue}{$Arg_2$} as correct and the counterfactual \textcolor{orange}{$Arg_2'$} as incorrect, and compute $\Delta L = L(Arg_2) - L(Arg_2')$, where $L(\cdot)$ denotes the logit of the corresponding answer.\\
\textbf{Normalized faithfulness:} Since different discourse relations yield different raw scores, we  report normalized faithfulness scores \cite{miller2024transformer}, which quantify the percentage of the full model’s performance that a sparse circuit restores. Concretely, we compute $\frac{\Delta L_{\mathrm{patch}}}{\Delta L_{\mathrm{full}}}$, where $\Delta L_{\mathrm{patch}}$ is the logit difference obtained by patching clean activations into a corrupted input, and $\Delta L_{\mathrm{full}}$ is the logit‐difference of the full model on clean input. In our \cudr task, faithfulness begins at a large negative value (since the unpatched model selects \textcolor{orange}{$Arg_2'$}), increases as clean edges are patched, and reaches 100\% when the full model is restored (which predicts \textcolor{blue}{$Arg_2$}).

\paragraph{Hierarchical Discursive Circuits.}

\begin{table}[h]
\footnotesize
\centering
\renewcommand{\arraystretch}{1.2} 
\resizebox{\columnwidth}{!}{%
\begin{tabular}{l c l}
\textbf{\LOne} & \textbf{\LTwo} & \textbf{\LThree} \\
\cline{1-3}
\multirow{2}{*}{\textbf{Comparison (568)} }  & \textcolor{gray}{Concession \ding{55}} & Arg2-as-denier \\
                                     \cline{2-3}
                                     & /                                      & Contrast \\
\cline{1-3}
\multirow{2}{*}{\textbf{Contingency (564)}} & /                                    & Reason \\
                                     \cline{2-3}
                                     & /                                     & Result \\
\cline{1-3}
\multirow{6}{*}{\textbf{Expansion (200)}}   & /                                    & Conjunction \\
                                     \cline{2-3}
                                     & /                                    & Equivalence \\
                                     \cline{2-3}
                                     & \textcolor{gray}{Instantiation \ding{55}} & Arg2-as-instance \\
                                     \cline{2-3}
                                     & \multirow{2}{*}{\textcolor{darkyellow}{Level-of-detail (565)} \checkmark} & Arg1-as-detail \\
                                     \cline{3-3}
                                     &                                          & Arg2-as-detail \\
                                     \cline{2-3}
                                     & \textcolor{gray}{Substitution \ding{55}}  & Arg2-as-subst \\
\cline{1-3}
\multirow{3}{*}{\textbf{Temporal (405)}}    & \multirow{2}{*}{\textcolor{darkyellow}{Asynchronous (575)} \checkmark} & Precedence \\
                                     \cline{3-3}
                                     &                                          & Succession \\
                                     \cline{2-3}
                                     & /                                       & Synchronous \\
\cline{1-3}
\end{tabular}
}
\caption{\textbf{Discursive Circuits Hierarchy (L1–L3):} All ``leaf node'' relations are classified as \LThree. Only two circuits appear at the \LTwo level, each merging more than one \LThree circuit. (Numbers) indicate edge counts. \LThree circuit has 1,000 edges, and \LZero circuit has 137 edges.}
\label{tab:hierarchical-relations}
\end{table}

With the learned circuits, we construct a new PDTB-style circuit hierarchy. To the best of our knowledge, this is the first discourse hierarchy derived from neural components.
We first learn circuits for all 13 Level-3 (\LThree) relations and use the top 1,000 edges to merge them to form higher-level circuits. That is, \LThree $\ni$ \LTwo $\ni$ \LOne $\ni$ \LZero (Table~\ref{tab:hierarchical-relations}). Note that our circuit hierarchy differs from the PDTB taxonomy in two ways: (1) All ``leaf node'' relations are treated as \LThree since they have no children to merge (e.g., Temporal.Synchronous)
and circuit discovery operates on the finest-grain level; 
(2) Some \LTwo relations are removed (e.g., Concession \ding{55}) as they contain only one valid \LThree relation due to data scarcity, so merging would be meaningless. In the end, \LTwo circuits contain over 500 edges, \LOne circuits have 200–500+ edges, and the meta \LZero circuit contains 137 edges.

\subsection{Discursive Circuits are Faithful (\textbf{RQ1})}

\label{sec: main exp}

\begin{figure}[t!]
    \centering
    \includegraphics[width=\columnwidth]{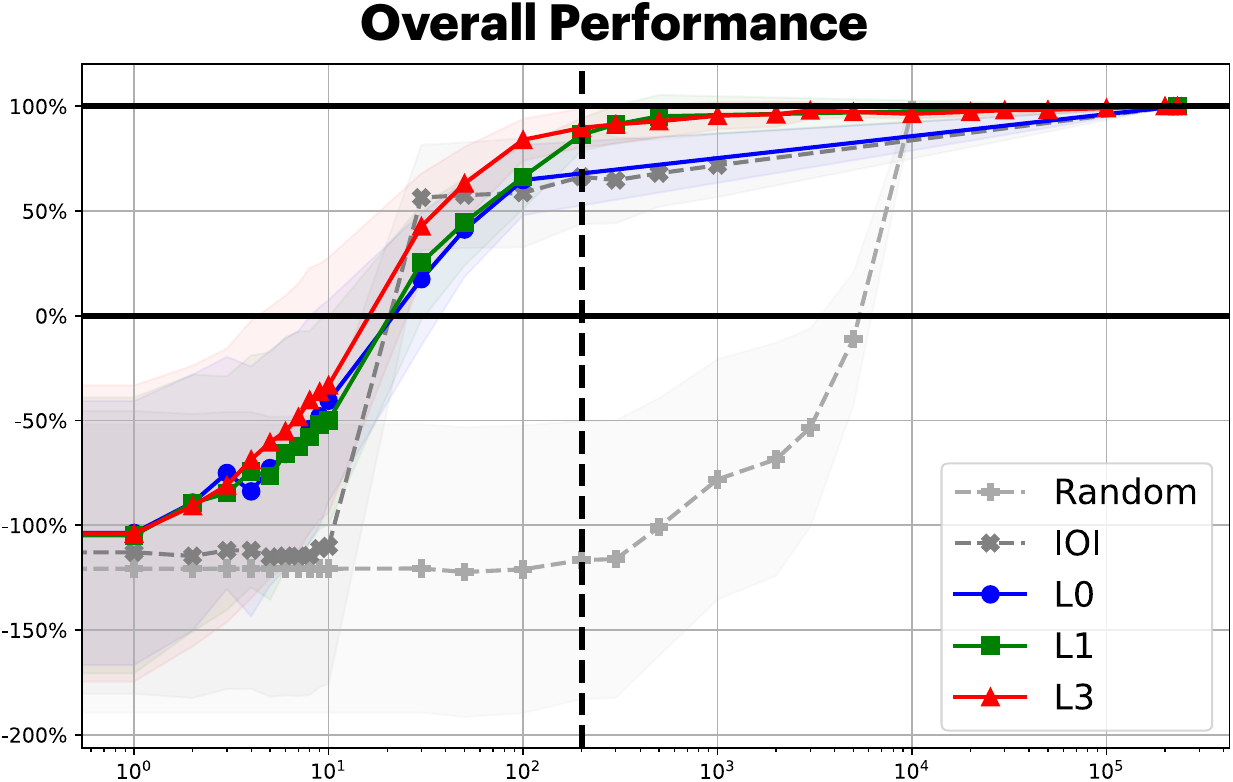}
    \caption{\textbf{RQ1: Overall Faithfulness of Discursive Circuits:} We report average faithfulness across 13 PDTB relations for circuits \LThree, \LOne, \LZero, the \random baseline, and the \IOI baseline. The Y-axis shows faithfulness (\%), and the X-axis shows the number of patched edges (log scale). Shaded areas indicate standard deviation. \LThree and \LOne reach strong faithfulness at $\approx 200$ edges (vertical dashed line).}
    \label{fig:overall_performance}
\end{figure}

\begin{figure}[h]
    \centering
    \includegraphics[width=7.6cm]{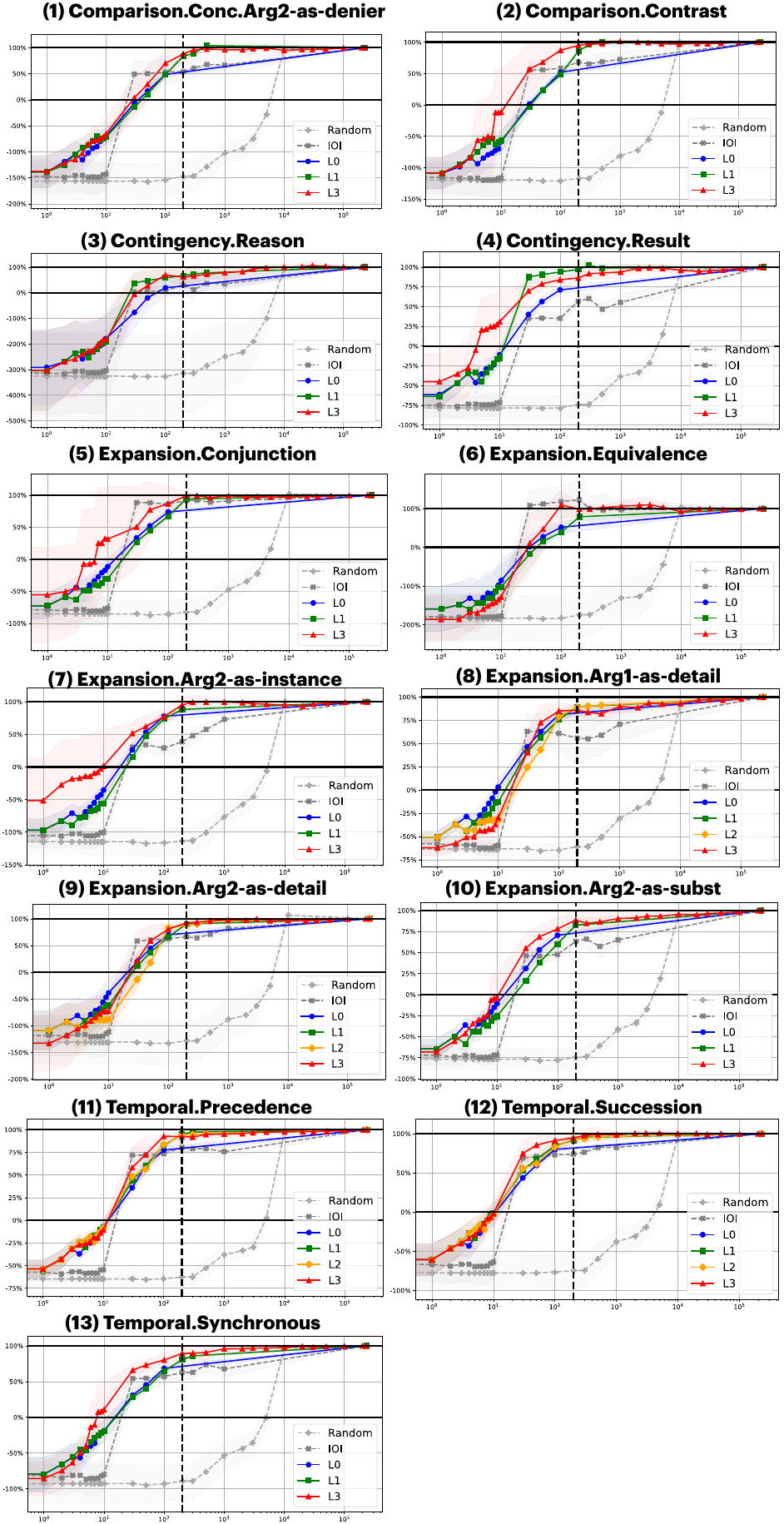}
    \caption{\textbf{RQ1} Faithfulness of Discursive Circuits by Discourse Relation (see indices 1–13). }
    \label{fig:seperate_performance}
\end{figure}
 
We first validate the faithfulness of discursive circuits on the PDTB dataset. The average performance across 13 discourse relations (Figure~\ref{fig:overall_performance}) shows strong overall effectiveness. We omit \LTwo as it covers only a subset of relations. 
For both \LThree and \LOne circuits, strong faithfulness ($\approx 90\%$) is achieved with only $\approx 200$ edges. \LThree outperforms \LOne in the 10–200 edge range, likely due to its ability to capture more fine-grained information. Both \LThree and \LOne surpass \LZero, \IOI, and \random after 100 edges. This gap is likely due to \LZero's small size (137 edges) which concentrates only on the most dominant skill.
The \random baseline shows almost no capacity to solve the \cudr task with fewer than 200 edges, and only begins to improve after 1,000 edges, indicating that the task requires non-trivial circuit structure to succeed.
Even though \IOI reasons over next objects, it still lacks discourse skills, as it plateaus quickly around $\approx 50\%$ faithfulness, showing the unique skills needed for discourse competence. 

We then analyze the performance breakdown by relation types (Figure~\ref{fig:seperate_performance}) and make the following observations:
\textbf{(1) Finer-grained circuits are more effective than coarser ones.} There is a consistent trend across relation types: \LThree\ $>$ \LTwo\ $\approx$ \LOne\ $>$ \LZero\ $>$ \IOI. However, fine-grained circuits also show greater variance (large \textcolor{red}{red shades}).
\LOne is more stable and has a lower variance. 
In practice, we recommend \LOne as a balanced choice: while slightly less effective in early stages, it matches \LThree after $\approx 300$ edges and works for all lower-level relations.
\textbf{(2) \LTwo does not necessarily outperform \LOne.} This is evident in the four relations that have \LTwo circuits, including Expansion.Details (8th and 9th subfigures in Figure~\ref{fig:seperate_performance}, compared with Expansion \LOne's circuit) and Temporal.Asynchronous (12th and 13th, compared with Temporal \LOne circuit). This suggests that \LTwo and \LOne operate at a similar level of abstraction, with comparable degrees of information loss.
\textbf{(3) Discursive circuits reflect task difficulty.} 
Two Contingency relations (3rd and 4th) are exceptions where \LOne matches or outperforms \LThree. Further inspection shows that these relations have lower absolute faithfulness scores, suggesting the model struggles with them. In such cases, \LThree may overfit, while \LOne retains core patterns and generalizes better.
\IOI generally underperforms due to its lack of discourse specificity. However, in Conjunction (5th) and Equivalence (6th), it performs comparably or better than discursive circuits, suggesting these relations are easier to model. In contrast, larger gaps in Comparison (1st–2nd) and Contingency (3rd–4th) indicate greater complexity.

\subsection{Discursive Circuits Generalize to New Datasets and New Relations (\textbf{RQ2})}

\label{sec: circuit generalization}
\begin{figure}[t!]
    \centering
    \includegraphics[width=7.6cm]{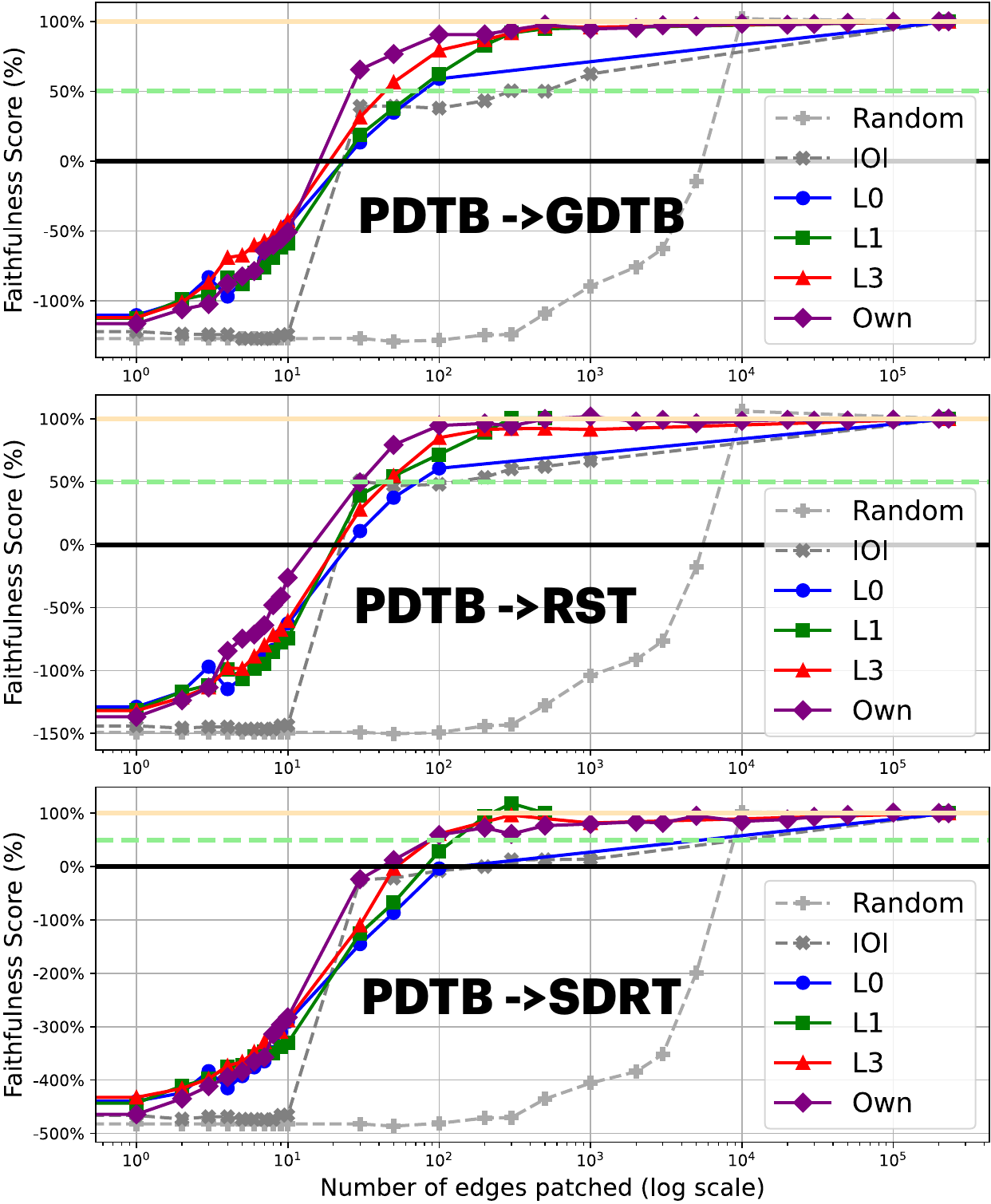}
    \caption{
    \textbf{RQ2 Cross-dataset generalization:} 
    Performance by applying PDTB's circuits to other datasets.
    }
    \label{fig:framework_gen}
\end{figure}

\newcommand*\circled[1]{\tikz[baseline=(char.base)]{
    \node[draw,circle,inner sep=0.1pt, line width=0.2pt,
    minimum size=0.9em, font=\footnotesize] (char) {#1};}}

\begin{figure}[t!]
    \centering
    \includegraphics[width=7.6cm]{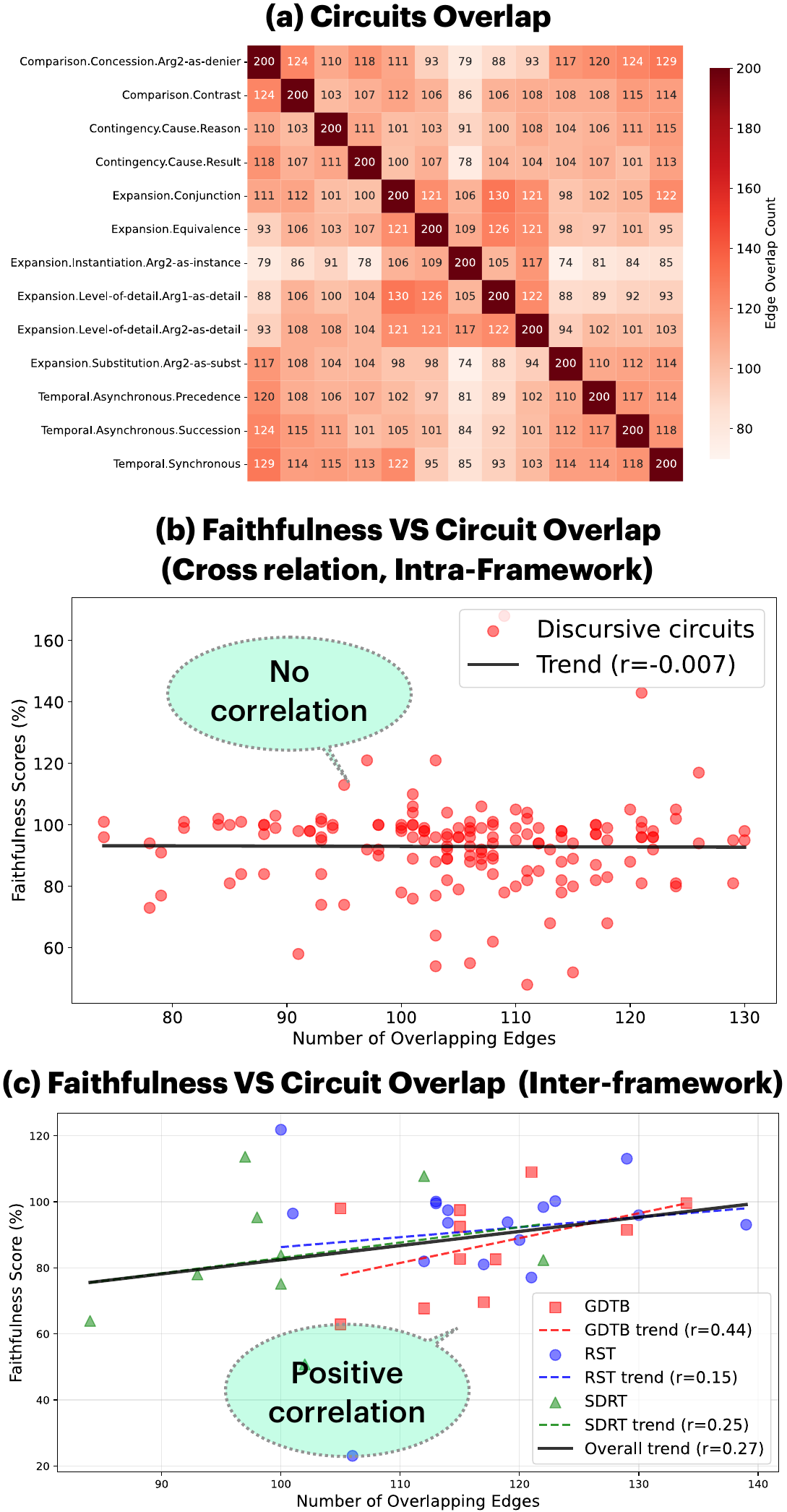}
    \caption{\textbf{RQ2 Cross-relation Generalization:} (a) The overlap among PDTB's relation circuits; (b) Intra-framework generalization in PDTB; (c) Inter‐framework generalization from PDTB.
}
    \label{fig:rel_gen}
\end{figure}

\textit{Do discursive circuits generalize across different discourse frameworks?} We extend the \cudr task to other frameworks by applying circuits obtained from PDTB to GDTB (same framework, different genre), as well as to RST and SDRT (different frameworks). We follow the same mapping (Appendix~\ref{app: dataset mappings}) for cross-framework transfer; for example, Explanation (SDRT) is mapped to Contingency.Cause.Reason (PDTB). Figure~\ref{fig:framework_gen} shows the generalization performance, with each line representing the average performance across all relations in the dataset.
\textbf{PDTB circuits generalize well to other datasets.} We set an ``upper bound'' using the \LOwn circuits (learned via \cudr task in-dataset, e.g. SDRT's Explanation). PDTB's \LThree circuits close the gap with \LOwn using only $\approx 200$ edges, despite initially lagging due to dataset-specific features. Across the three generalization targets, we observe \LOwn $>$ \LThree $>$ \LOne $\approx$ \LZero $>$ \IOI $>$ \Random, which is a consistent trend. \LOne and \LZero are weaker in the first 100 edges, likely because both abstractions lose fine-grained information (\LTwo is skipped due to limited coverage). SDRT is the most challenging to generalize to, with only 50\% faithfulness after 100 patched edges, highlighting the gap between the datasets.

\textit{Do circuits learned for one discourse relation generalize to others?}
We study all 13 PDTB \LThree relations by applying each circuit to the other 12, using the top 200 edges per circuit (enough for strong faithfulness):
\textbf{(1)} Figure~\ref{fig:rel_gen}a shows the edge overlap among these circuits. While the diagonals are darker, indicating greater overlap between similar relations, the overall overlap remains consistently high (80–120 out of 200 edges).
\textbf{(2)} Figure~\ref{fig:rel_gen}b shows no correlation between overlap and faithfulness ($r = -0.007$). This is counterintuitive, as one might expect higher overlap to imply better generalization. The narrow overlap range (80–120) likely limits the variation. Recently, \citet{hanna2024have} also reports faithfulness does not necessarily require high overlap.
\textbf{(3)} Cross-framework results (Figure~\ref{fig:rel_gen}c) reveal a positive correlation between overlap and performance, e.g., PDTB $\rightarrow$ GDTB yields $r = 0.44$.
In summary, higher circuit overlap \textit{does not} imply better intra-framework faithfulness, but \textit{does} support inter-framework transfer.

\subsection{Discursive Circuits Overlap with Linguistic Features' Circuits (\textbf{RQ3})}

\label{sec: linguistic features}

\begin{figure*}[ht!]
    \centering
    
    \begin{subfigure}{\textwidth}
        \centering
        \includegraphics[width=16cm]{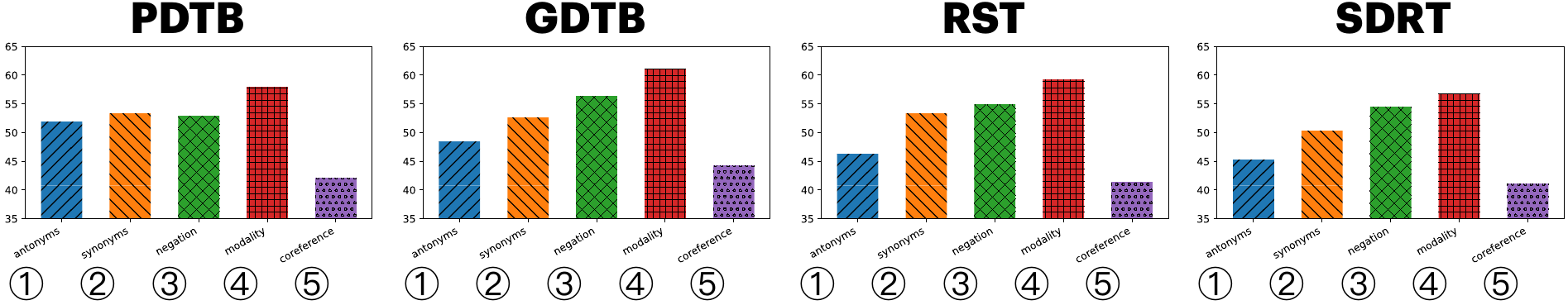}
        \caption{\textbf{Average overlap} between discursive circuits and circuits for linguistic features (averaged over all discourse relations within a framework). A consistent trend shares across frameworks: \textcolor{MyRed}{4.modality} is most heavily utilized, while \textcolor{MyPurple}{5.coreference} is the least.
        }
        \label{fig:linguistic_features_a}
    \end{subfigure}
    
    \vspace{0.2cm}
    
    \begin{subfigure}{\textwidth}
        \centering
        \includegraphics[width=16cm]{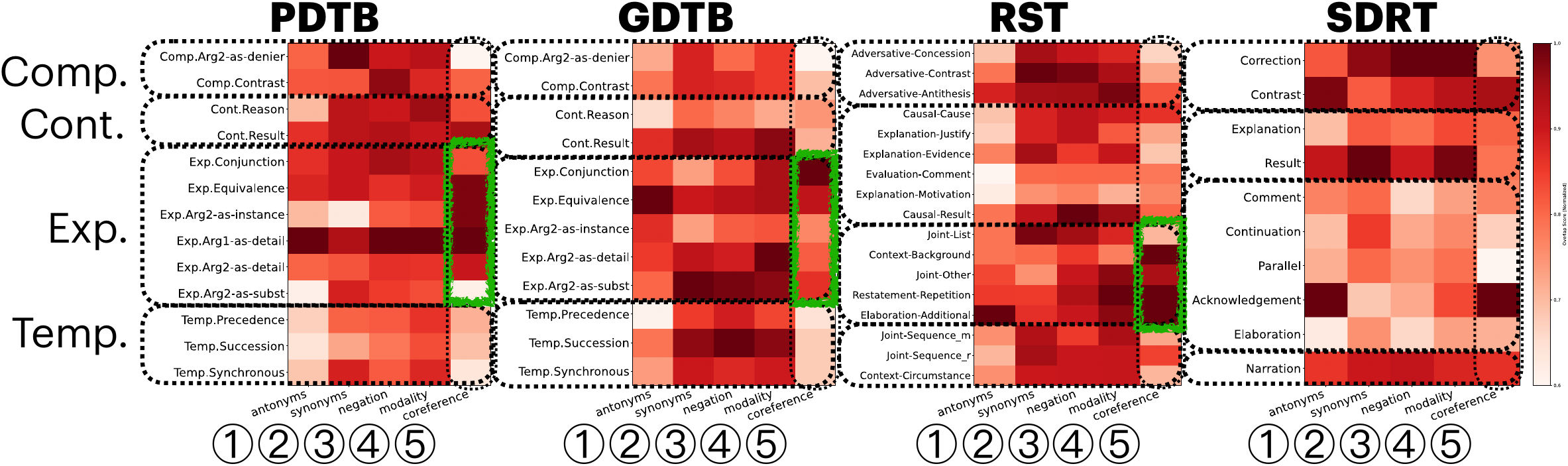}
        \caption{\textbf{Normalized overlap} (column-wise), where each column is scaled such that its maximum value equals 1 (values in heatmaps range from 0.6 to 1). Similar cross-framework patterns are observed, for example, modality is strongly utilized across all frameworks, while \textcolor{darkgreen}{coreference signals} appear prominently in most Expansion relations.}
        \label{fig:linguistic_features_b}
    \end{subfigure}

    \caption{\textbf{RQ3} Overlap of discursive circuits with circuits for linguistic features: antonymy, synonymy, negation, modality, and coreference.}
    \label{fig:linguistic_features}
\end{figure*}

\textit{Are discursive circuits composed of sub-circuits linked to linguistic features?} Inspired by the eRST and RST Signaling Corpus \cite{zeldes-etal-2025-erst, das2018rst}, we discover circuits for five key features, \circled{1} antonymy, \circled{2} synonymy, \circled{3} negation, \circled{4} modality, and \circled{5} coreference, as a preliminary and non-exhaustive study,
using similar activation prompts (Appendix~\ref{app: detail experiments}).
We find that the utilities of linguistic features are broadly consistent across frameworks (Figure~\ref{fig:linguistic_features_a}). Utility is measured as the overlap between circuits associated with a given linguistic feature and the discovered discursive circuits, averaged over all discourse relations within a framework. Among these features, \circled{4} modality is the most extensively utilized, while \circled{5} coreference is the least. Interestingly, the \circled{2} synonymy feature is consistently more prominent than \circled{1} antonymy across all frameworks, suggesting that synonymy serves as a more common cohesive device. We also find that irrelevant circuits overlap only weakly with discursive circuits (e.g., IOI overlaps with PDTB circuits on only about 20 edges).
To enable a fair, fine-grained comparison across linguistic features, we present column-wise normalized overlaps (Figure~\ref{fig:linguistic_features_b}). Normalization ensures that each feature is scaled relative to its own maximum, allowing comparison across frameworks without one feature dominating due to raw magnitude. We find a consistent utility at the level of individual discourse relations. From a broad perspective, PDTB, GDTB, and RST display similar heatmap structures, while SDRT diverges significantly. Across the three similar frameworks, \circled{2} synonymy, \circled{3} negation, and \circled{4} modality are heavily used across most relations. In contrast, \circled{1} antonymy is relatively weak in Contingency and Temporal relations (lighter-colored cells). Notably, \circled{5} coreference is most active in Expansion relations (highlighted by the darkest \circled{5} cells within the \textbf{\textcolor{darkgreen}{green boxes}}), reflecting the role of entity continuity. SDRT, however, shows less reliance on coreference, likely due to shorter texts.

\begin{figure}[ht!]
    \centering
   \includegraphics[width=\columnwidth]{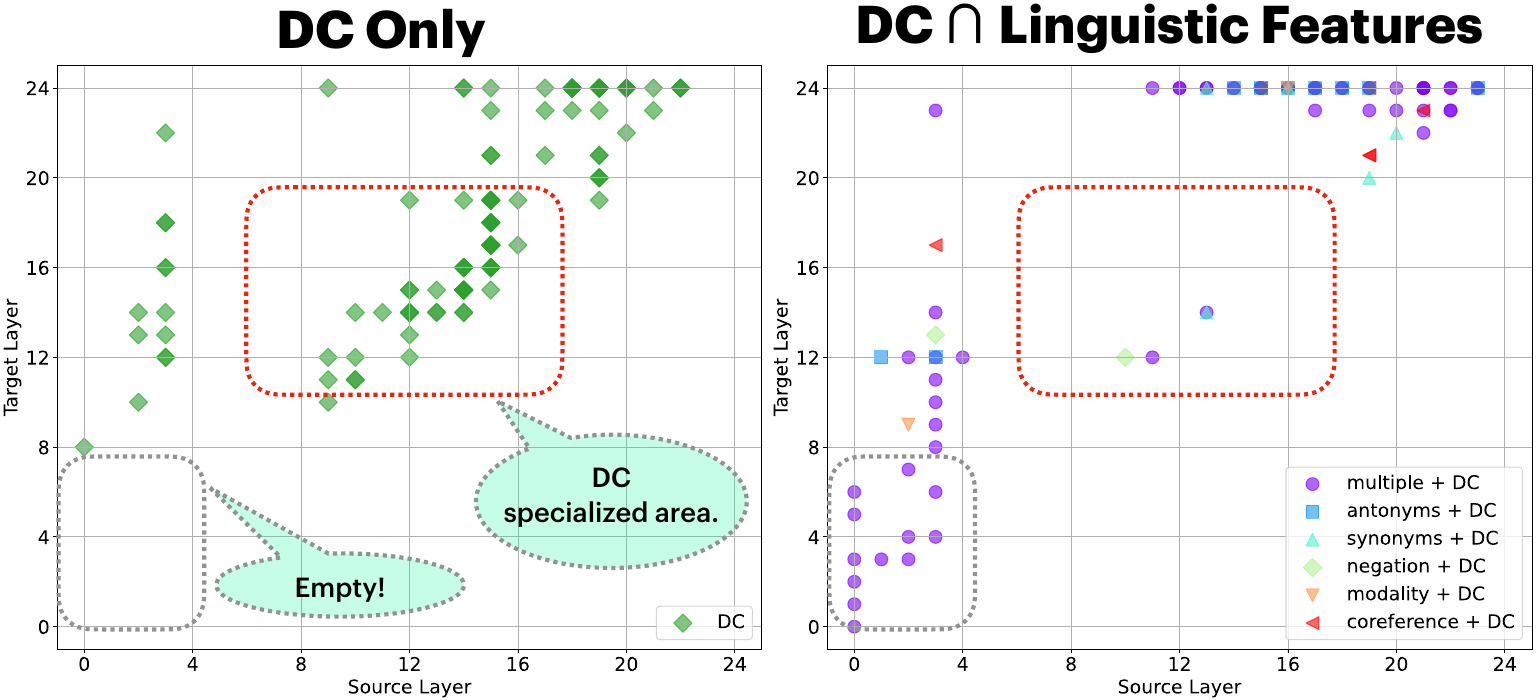}
    \caption{ \textbf{RQ3 Layer-wise Edge Analysis:} Source (X-axis) and target (Y-axis) layers of edges in discursive and linguistic circuits. DC-only edges emerge in higher layers and are absent in lower layers.
    }
    \label{fig:source_target}
\end{figure}
Figure~\ref{fig:source_target} shows the layer-wise distribution of discursive circuits (DC) and linguistic circuits by source and target node layers (Top 200 edges). DC-only edges are absent in lower layers (noted as ``empty''). A distinct region (source: 8–16, target: 10–20) contains DC-only edges, with very limited overlap with linguistic features. This suggests lower layers in discursive circuits capture shared linguistic features, while discursive abstraction emerges in higher layers.

\begin{table}[t!]
\small
\begin{tabular}{|p{7.2cm}|}
\hline
\textbf{Error case 1:} [yay!!!!!]$_{Arg_1}$, (because) [I don't care who wins now]$_{Arg_2}$\\
\textbf{Error case 2:} [I'll give clay in return]$_{Arg_1}$, (because) [think clay is in abundance this game]$_{Arg_2}$\\
\textbf{PDTB's missing edges:} Resid Start$\rightarrow$MLP0, A19.9$\rightarrow$A21.1, MLP3$\rightarrow$MLP7, MLP7$\rightarrow$MLP11\\
 \hline
\end{tabular}
\caption{\textbf{Case Study:} PDTB circuit \ding{55}; SDRT circuit \checkmark}
\label{tab:case study}
\end{table}

We further examine the cases where SDRT's \LOwn circuits succeed but PDTB's \LThree circuits fail (both using the first 30 edges). Table \ref{tab:case study} shows a subset of representative errors. Case 1 involves an interjection (``yay!''), and Case 2 features an ellipsis of the subject ``I'' in $Arg_2$, both are rare phenomena in PDTB. Our method pinpoints missing elements in PDTB that SDRT captures, such as early edges (Resid Start$\rightarrow$MLP 0, aiding connective reasoning) and late edges (e.g., 19.9$\rightarrow$21.1, shared only with the coreference feature among the five features).

\section{Related Works}
\paragraph{Discourse Modeling and Evaluation.} 
Discourse modeling has been studied under three major frameworks: PDTB \cite{webber2019penn, prasad2008penn}, RST \cite{mann1987rhetorical, zeldes2017gum, zeldes-etal-2025-erst}, and SDRT \cite{asher2003logics}. 
Recent studies seek to unify these frameworks, with advances in discourse relation prediction \cite{zhao-etal-2023-infusing, wu-etal-2023-connective, anuranjana-2023-discoflan, chan-etal-2023-discoprompt, liu-strube-2023-annotation, rong-mo-2024-ncprompt, li-etal-2024-discourse, liu2025jointmodelingentitiesdiscourse, long-etal-2024-multi, aktas-roth-2025-clarifying}, 
discourse structure parsing \cite{li-etal-2023-discourse, li-etal-2024-dialogue, thompson-etal-2024-llamipa, liu-etal-2025-enhancing, zhang-etal-2025-structured, namuduri2025qudsim}, and annotation \cite{pyatkin-etal-2023-design, yung-etal-2024-prompting, ruby-etal-2025-multimodal, saeed-etal-2025-implicit}.  \citet{fu-2022-towards} outlines early plans for unification, 
and the DISRPT benchmark \cite{braud-etal-2024-disrpt} enables cross-framework evaluation with data annotated under all three schemes. \citet{liu-etal-2024-gdtb} propose automatic RST-to-PDTB transformation via sense mapping. \citet{liu-zeldes-2023-cant, eichin2025probingllmsmultilingualdiscourse} examine generalization across domains and languages. While linguistically insightful, existing approaches overlook interpretability.

Question answering has also been explored as a bridge across frameworks. \citet{fu2025surveyqudmodelsdiscourse} links Questions Under Discussion (QUD) \cite{wu2023qudeval, ko-etal-2023-discourse} to PDTB, RST, and SDRT. \citet{miao-etal-2024-discursive} propose a QA-based evaluation, though their prompts offer limited insight into model internals.
LLMs have been used to synthesize discourse data \cite{yung2025syntheticdataaugmentationcrossdomain, cai2025finegrainedevaluationimplicitdiscourse}, mainly to augment low-resource relations \cite{omura-etal-2024-empirical}. In contrast, our \cudr dataset targets interpretability rather than data expansion.

\paragraph{Mechanistic Interpretability.}
Unlike visualizations \cite{jain-wallace-2019-attention, wiegreffe-pinter-2019-attention} or textual explanations \cite{lyu-etal-2024-towards, zhu-etal-2024-explanation}, mechanistic interpretability 
identifies components in a model that drive predictions. Circuits, as global computation graphs, can be identified through activation patching \cite{conmy2023towards, miller2024transformer, syed-etal-2024-attribution, bakalova2025contextualizethenaggregate}. We do not adopt sparse autoencoders (SAEs) \cite{huben2024sparse, makelov2024towards} or neuron-level analysis \cite{dai-etal-2022-knowledge,ai2025are}, as our goal is to understand discourse processing at a global model rather than isolate local activity. Circuit discovery has mostly been applied to simplistic tasks, such as indirect object identification (IOI) \cite{wang2023interpretability}, numerical comparison \cite{hanna2023how}, subject-verb agreement (SVA) \cite{ferrando-costa-jussa-2024-similarity}, MCQ \cite{lieberum2023does}, knowledge acquisition \cite{yao2024knowledge, ou2025llmsacquirenewknowledge, hanna2024have}, colored objects \cite{merullo2024circuit}, extractive QA \cite{basu2025on}, and context-free grammars \cite{mondorf2025circuitcomposition}. 
No existing work addresses complex discourse phenomena. 

\section{Conclusion and Future Work}
In this work, we introduce discursive circuits, the first mechanistic interpretation of how discourse understanding is realized within language models. To make circuit discovery feasible, we propose a novel \cudr task that enables activation patching, along with a collection of \cudr datasets for PDTB, RST, and SDRT discourse frameworks.
Our identified discursive circuits are shown to be faithful in restoring the full model's performance and exhibit strong cross-framework generalization. 
Discursive circuits provide a new lens for mechanistically representing discourse, enabling the construction of a circuit hierarchy that supports direct comparison of discourse relations both within and across frameworks. Based on that, we observe shared linguistic feature utility across frameworks. 
In future work, we plan to extend \cudr to diverse discourse styles and languages, and adapt it to broader tasks such as steering models in biased contexts and guiding future discourse taxonomy development.

\section*{Limitations}
Our work also has the following limitations: (1) We only study English-based corpora. It would be promising to extend circuit discovery to multiple languages and explore whether a unified circuit space exists across different languages, similar to the universal discourse label set explored by \citet{eichin2025probingllmsmultilingualdiscourse}. This is feasible, as we can construct the \cudr dataset for other languages as well.
(2) We follow \citet{hanna2023how, hanna2024have, mondorf2025circuitcomposition} in focusing on a single transformer-based language model to enable more in-depth analysis. While it would be interesting to extend our method to other model architectures such as multi-layer perceptrons (MLPs) \cite{fusco-etal-2023-pnlp} or LSTMs \cite{sundermeyer2012lstm}, we limit our scope to transformers due to their predominant use today and because activation patching is not directly compatible with MLPs or LSTMs.
(3) We do not compare discourse processing in language models with that in the human brain \cite{case2015brain, perfetti2008neural}. For example, \citet{eviatar2006brain} report that discourse processing triggers specific brain activations observable via fMRI. While intriguing, this is beyond the scope of our study.

\section*{Ethical Statement and Potential Risks}
Our research on discourse relations does not pose direct ethical risks. However, as with all mechanistic interpretability studies, the identified circuits could be used to influence model behavior in specific capacities, such as modifying numerical reasoning \cite{hanna2023how} or, in our case, discourse processing and generation. By making the model's reasoning about discourse relations more transparent, our work has the potential to aid in detecting and mitigating biases in scenarios where discourse structure plays a role.

The risk of data contamination in GPT-2 is low. Trained on the ``WebText'' corpus (Reddit-linked contents), GPT-2 explicitly excludes paywalled sources such as the Wall Street Journal, making inclusion of the PDTB corpus unlikely. The GUM corpus (GDTB, RST) comprises small, academically curated texts unlikely to appear in WebText, while the SDRT (STAC corpus) consists of annotated Catan dialogue logs, also absent from typical pretraining data.

\section*{Declaration of AI Tool Usage}
We used AI tools at the following stages of this research: (1) GPT-4o-mini (via API) was used to generate the counterfactual instances for our \cudr dataset; prompt details are provided in Appendix~\ref{app: dataset}; (2) Cursor AI was used during coding, primarily for debugging assistance; (3) ChatGPT-4o (via web interface) was employed only for grammar checking of the manuscript. All research ideas, analyses, and findings were developed and written independently by the authors.

\section*{Acknowledgements}
We thank our anonymous reviewers for their time spent on reviewing our paper and their detailed feedback, which greatly helped us refine our work.
We also thank several colleagues at National University of Singapore (NUS) for research discussions and proofreading of our drafts, especially Barid Xi Ai, Shumin Deng, Yajing Yang, Tongyao Zhu, Mahardika Krisna Ihsani, Xuan Long Do, and Xinyuan Lu. We appreciate Joseph Miller, Bilal Chughtai, and William Saunders for open sourcing their software \footnote{\url{https://ufo-101.github.io/auto-circuit/}} and Neel Nanda's blog \footnote{\url{https://www.alignmentforum.org/users/neel-nanda-1}} that guided the first author into mechanistic interpretability. 
We would also like to acknowledge a grant from National Research Foundation, Singapore under its AI Singapore Programme (AISG Award No: AISG2-GC-2022-005).

\bibliography{custom}

\appendix

\clearpage
\newpage
\appendix

\section{\cudr Dataset Details}
\label{app: dataset}

\subsection{Counterfactual Connectives}
\label{app: cf conn}
To create counterfactual instances in the \cudr dataset, we rely on the taxonomy by \citet{miao-etal-2024-discursive}, which defines each discourse relation along with five irrelevant counterfactual relations. Due to space constraints, Table~\ref{tab:conn} in Section~\ref{sec: CuDR} lists only a subset of the counterfactual connectives. The complete set of five counterfactual connectives is provided in Table~\ref{tab:conn_full}.

\begin{table}[h]
\footnotesize
\centering
\resizebox{\columnwidth}{!}{%
\begin{tabular}{lll}
\textbf{Discourse Relation} & \textbf{Ori Connective} & \textbf{CF Connectives} \\ 
\hline
\textbf{Comparison.Concession.Arg2-as-denier} 
  & however 
  & \makecell[l]{because \\ for example \\ specifically \\ so \\ in other words} \\
\hline
\textbf{Comparison.Contrast} 
  & by comparison 
  & \makecell[l]{specifically \\ in other words \\ because \\ for example \\ so} \\
\hline
\textbf{Contingency.Reason} 
  & because 
  & \makecell[l]{so \\ however \\ by comparison \\ for example \\ in other words} \\
\hline
\textbf{Contingency.Result} 
  & so 
  & \makecell[l]{because \\ by comparison \\ for example \\ however \\ in other words} \\
\hline
\textbf{Expansion.Conjunction} 
  & and 
  & \makecell[l]{however \\ so \\ because \\ by comparison \\ instead} \\
\hline
\textbf{Expansion.Equivalence} 
  & in other words 
  & \makecell[l]{however \\ for example \\ because \\ so \\ by comparison} \\
\hline
\textbf{Expansion.Instantiation.Arg2-as-instance} 
  & for example 
  & \makecell[l]{because \\ however \\ by comparison \\ so \\ in other words} \\
\hline
\textbf{Expansion.Level-of-detail.Arg1-as-detail} 
  & in short 
  & \makecell[l]{however \\ so \\ by comparison \\ in other words \\ instead} \\  
\hline
\textbf{Expansion.Level-of-detail.Arg2-as-detail} 
  & specifically 
  & \makecell[l]{instead \\ by comparison \\ however \\ so \\ in other words} \\
\hline
\textbf{Expansion.Substitution.Arg2-as-subst} 
  & instead 
  & \makecell[l]{because \\ in other words \\ so \\ for example \\ specifically} \\
\hline
\textbf{Temporal.Asynchronous.Precedence} 
  & then 
  & \makecell[l]{however \\ previously \\ by comparison \\ for example \\ because} \\
\hline
\textbf{Temporal.Asynchronous.Succession} 
  & previously 
  & \makecell[l]{so \\ then \\ by comparison \\ however \\ for example} \\
\hline
\textbf{Temporal.Synchronous} 
  & while 
  & \makecell[l]{so \\ then \\ by comparison \\ however \\ for example} \\
\hline
\end{tabular}%
}
\caption{\textbf{\cudr Dataset Details (Full Counterfactual Connectives):} PDTB discourse relations with their original (Ori) connective and the corresponding set of five counterfactual (CF) connectives.}
\label{tab:conn_full}
\end{table}

\subsection{Aligning Discourse Frameworks}
\label{app: dataset mappings}

We refer to cross-framework relation mapping both to prepare counterfactual \cudr data for frameworks beyond PDTB (Section~\ref{sec: cudr datasets}) and to perform cross-framework transfer (Section~\ref{sec: circuit generalization}). The mapping between PDTB and the GUM Discourse Treebank (GDTB) \cite{liu-etal-2024-gdtb} is straightforward, as GDTB adopts the PDTB relation taxonomy.
For the GUM Rhetorical Structure Theory (GUM-RST) dataset \cite{zeldes2017gum}, we closely examine the annotation guidelines and the mapping approach used by \citet{liu-etal-2024-gdtb}. Based on this, we define a mapping shown in Table~\ref{tab:rst2pdtb}, which includes 17 RST relations, excluding those with insufficient data. This mapping offers broad coverage, aligning the 17 RST relations with 9 distinct PDTB relations.
For the Segmented Discourse Representation Theory (SDRT) dataset \cite{asher2003logics}, we also examine the relation definitions and construct the mapping presented in Table~\ref{tab:sdrt2pdtb}. This results in 10 distinct SDRT relations mapped to 8 PDTB relations.

\begin{table}[h]
\footnotesize
\centering
\setlength{\tabcolsep}{6pt}
\renewcommand{\arraystretch}{1.2}
\resizebox{\columnwidth}{!}{%
\begin{tabular}{ll}

\textbf{RST Label} & \textbf{Mapped PDTB Label} \\
\hline
\textbf{joint-list\_m} & Expansion.Conjunction \\
\textbf{joint-sequence\_m} & Temporal.Asynchronous.Precedence \\
\textbf{elaboration-additional\_r} & Expansion.Level-of-detail.Arg2-as-detail \\
\textbf{context-circumstance\_r} & Temporal.Synchronous \\
\textbf{adversative-concession\_r} & Comparison.Concession.Arg2-as-denier \\
\textbf{causal-cause\_r} & Contingency.Cause.Reason \\
\textbf{causal-result\_r} & Contingency.Cause.Result \\
\textbf{adversative-contrast\_m} & Comparison.Contrast \\
\textbf{explanation-justify\_r} & Contingency.Cause.Reason \\
\textbf{context-background\_r} & Expansion.Conjunction \\
\textbf{joint-other\_m} & Expansion.Conjunction \\
\textbf{adversative-antithesis\_r} & Comparison.Contrast \\
\textbf{explanation-evidence\_r} & Contingency.Cause.Reason \\
\textbf{evaluation-comment\_r} & Contingency.Cause.Reason \\
\textbf{explanation-motivation\_r} & Contingency.Cause.Reason \\
\textbf{restatement-repetition\_m} & Expansion.Equivalence \\
\textbf{joint-sequence\_r} & Temporal.Asynchronous.Precedence \\
\hline
\end{tabular}%
}
\caption{\textbf{RST to PDTB Mapping:} Mapping of RST discourse labels to PDTB labels for the \cudr dataset.}
\label{tab:rst2pdtb}
\end{table}

\begin{table}[h]
\footnotesize
\centering
\setlength{\tabcolsep}{6pt}
\renewcommand{\arraystretch}{1.2}
\resizebox{\columnwidth}{!}{%
\begin{tabular}{ll}

\textbf{SDRT Label} & \textbf{Mapped PDTB Label} \\
\hline
\textbf{Acknowledgement} & Expansion.Equivalence \\
\textbf{Comment}         & Expansion.Conjunction \\
\textbf{Continuation}    & Expansion.Conjunction \\
\textbf{Contrast}        & Comparison.Contrast \\
\textbf{Correction}      & Comparison.Concession.Arg2-as-denier \\
\textbf{Elaboration}     & Expansion.Level-of-detail.Arg2-as-detail \\
\textbf{Explanation}     & Contingency.Cause.Reason \\
\textbf{Narration}       & Temporal.Asynchronous.Precedence \\
\textbf{Parallel}        & Expansion.Conjunction \\
\textbf{Result}          & Contingency.Cause.Result \\
\hline
\end{tabular}%
}
\caption{\textbf{SDRT to PDTB Mapping:} Mapping of SDRT discourse labels to PDTB labels for the \cudr dataset.}
\label{tab:sdrt2pdtb}
\end{table}

\subsection{Details for \cudr Dataset Construction}
\label{app: dataset construction}

To construct the counterfactual argument \textcolor{orange}{$Arg'_2$}, we ensure it is coherent with both the original argument \textcolor{blue}{$Arg_1$} and the counterfactual discourse relation, along with its connective \textcolor{orange}{$Conn'$}.
\textbf{Input:} We generate the dataset by prompting the GPT-4o-mini model via API, chosen for its balance of instruction-following ability and efficiency. Each prompt includes \textcolor{blue}{$Arg_1$}, \textcolor{orange}{$Conn'$}, and a \texttt{CF\_dr\_description} field defining the discourse relation. For example, \texttt{Contingency.Cause.Reason} is described as ``$Arg_2$ is the reason for $Arg_1$: when $Arg_1$ gives the effect, and $Arg_2$ provides the reason, explanation, or justification'', adapted from the PDTB annotation guidelines \cite{webber2019penn}.
\textbf{Requirements:} We ask the model to complete a structured JSON template. To maintain quality and discourage shallow completions, we explicitly instruct the model \textit{not} to repeat \textcolor{orange}{$Conn'$} verbatim, and instead to use relation-specific language patterns. We also request that \textcolor{orange}{$Arg'_2$} match the length of \textcolor{blue}{$Arg_1$}, improving stylistic and structural consistency.
\textbf{Output and Postprocessing:} The model is prompted independently for each \cudr data instance, and its output is saved as a plain text file. These files are subsequently parsed into usable JSON format using a custom loader.
The final prompt template, with inserted variables such as \textcolor{blue}{$Arg_1$} and \textcolor{orange}{$Conn'$}, is shown below:

\noindent\hspace{+0.15cm}\begin{minipage}{7.4cm}
\begin{lstlisting}[
  basicstyle=\ttfamily\scriptsize,
  breaklines=true,
  columns=fullflexible,
  backgroundcolor=\color{green!10},
  frame=single
]
You are an expert in discourse semantics. In discourse theory, arg1 and arg2 are two arguments connected by a relation (a connective word).
I am going to give you an original discourse argument (*original_arg1*) and a counterfactual relation (*CF_dr*). Your task is to generate a new counterfactual argument (*counterfactual_arg2*) that aligns with *original_arg1* while reflecting the given counterfactual relation.

**Requirements:**
1. *counterfactual_arg2* must be **coherent** with *original_arg1* and appropriately reflect the given counterfactual relation (by writing after counterfactual_connective).
2. The length of *counterfactual_arg2* should be around {original_arg2_length} words.
3. Make the relation between *counterfactual_arg2* and *original_arg1* easy to understand and as salient as possible.
4. Do not repeat the connective word in your *counterfactual_arg2*. Instead, try to use negation or contrastive signal (for comparison counterfactuals), specific causal events of result or reason (for contingency counterfactual), specific examples like entities and concrete details (for expansion counterfactuals).

Complete the following dictionary and only return the dictionary as your output:
{
    "original_arg1": "{original_arg1}",
    "counterfactual_relation": "{CF_dr}", which means {CF_dr_description},
    "counterfactual_connective": "{conn_CF}",
    "counterfactual_arg2": TO BE COMPLETED
}
\end{lstlisting}
\end{minipage}

\paragraph{Manual Verification}

One author manually verified the quality of our \cudr data samples. We randomly sampled 10 instances from each discourse framework and present subsets of \cudr examples from the PDTB (Table~\ref{tab:cudr_samples_pdtb}), GDTB (Table~\ref{tab:cudr_samples_gdtb}), RST (Table~\ref{tab:cudr_samples_rst}), and SDRT (Table~\ref{tab:cudr_samples_sdrt}) datasets.
Although each framework uses different terminology, we adopt a unified notation of $Arg_1$ and $Arg_2$ throughout.
\textbf{Across the 40 samples, we find all to be valid:} the generated \textcolor{orange}{$Arg'_2$} is coherent with the original \textcolor{blue}{$Arg_1$} and aligns well with the intended counterfactual connective \textcolor{orange}{$Conn'$}.
For example, in the first PDTB sample, the original \textcolor{blue}{$Arg_1$} is ``Robert S. Ehrlich resigned as chairman, president and chief executive'', which is linked by a denying relation (signaled by ``however'') to ``Mr. Ehrlich will continue as a director and a consultant''. Under the counterfactual connective \textcolor{orange}{$Conn'$} ``so'', our generated \textcolor{orange}{$Arg'_2$} becomes ``the company faced significant leadership challenges afterward'', directly expressing the consequence of Mr. Ehrlich’s resignation and appropriately realizing the intended relation.
Beyond PDTB, our \cudr construction performs well across other frameworks. For instance, although SDRT often contains shorter text spans, the generated \textcolor{orange}{$Arg'_2$} still effectively reflects the intended \textcolor{orange}{$Conn'$}. In Sample 2 from Table~\ref{tab:cudr_samples_sdrt}, ``others settle for less'' clearly presents a contrasting scenario, demonstrating that the model can express discourse relations concisely.

\textbf{However, we do find our generated data to be straightforward in their expression.} In all samples we examined, rare words are seldom used, and the model tends to prefer simple sentence structures. For example, Sample 3 in SDRT (Table~\ref{tab:cudr_samples_sdrt}) has an original \textcolor{blue}{$Arg_1$} as ``yep saturday's looking promising'', and continues with an \textcolor{orange}{$Arg'_2$} expression ``the weather forecast predicts sunshine'', using the counterfactual connective ``because''. This is a valid instance, but discussing the weather is relatively expected and less surprising. Sample 3 in PDTB (Table~\ref{tab:cudr_samples_pdtb}) has an \textcolor{blue}{$Arg_1$} as ``Much is being done in Colombia to fight the drug cartel mafia'', and it assigns \textcolor{orange}{$Arg'_2$} as ``the government recognizes that drug trafficking severely undermines national security and social stability''. While this is a valid continuation aligning with the counterfactual connective ``because'', it lacks specific knowledge about the drug situation in Colombia. In contrast, the original \textcolor{blue}{$Arg_2$} is ``luxurious homes and ranches have been raided by the military authorities, and sophisticated and powerful communications equipment have been seized'', which offers more phrase variation and concrete detail.

This straightforward style is expected, since we explicitly prompt the model: ``Make the relation between counterfactual $Arg'_2$ and original $Arg_1$ easy to understand and as salient as possible.'' We design the \cudr task this way to maximize its steering effect. Future work could extend \cudr to more complex texts and ambiguous scenarios, and explore controlling the attributes and style of $Arg'_2$ through in-context learning (ICL) \cite{liu-wang-2023-towards, long-etal-2025-beyond, li-etal-2025-delta}.

\begin{table}[h!]
\small
\begin{tabular}{|p{7.2cm}|}
\hline
\textbf{\cudr data samples for PDTB framework:} \\[0.5ex]

\textbf{Sample 1: } \\
\textcolor{blue}{$Arg_1$}: Robert S. Ehrlich resigned as chairman, president and chief executive \\
\textcolor{blue}{$Arg_2$}: Mr. Ehrlich will continue as a director and a consultant\\
\textcolor{blue}{Original relation}: Comparison.Concession.Arg2-as-denier (however) \\
\textcolor{orange}{$Conn'$}: so \\
\textcolor{orange}{$Arg'_2$}: the company faced significant leadership challenges afterward. \\ \hline

\textbf{Sample 2: } \\
\textcolor{blue}{$Arg_1$}: Shortly after Texas Air took control of Eastern, some Machinists union supervisors received a 20\% pay raise \\
\textcolor{blue}{$Arg_2$}: the pilots argued that this triggered a pay raise for them\\
\textcolor{blue}{Original relation}: Contingency.Cause.Result (so) \\
\textcolor{orange}{$Conn'$}: but \\
\textcolor{orange}{$Arg'_2$}: most other employees were not granted any wage increase. \\ \hline

\textbf{Sample 3: } \\
\textcolor{blue}{$Arg_1$}: Much is being done in Colombia to fight the drug cartel mafia\\
\textcolor{blue}{$Arg_2$}: luxurious homes and ranches have been raided by the military authorities, and sophisticated and powerful communications equipment have been seized\\
\textcolor{blue}{Original relation}: Expansion.Instantiation.Arg2-as-instance (for example) \\
\textcolor{orange}{$Conn'$}: because \\
\textcolor{orange}{$Arg'_2$}: the government recognizes that drug trafficking severely undermines national security and social stability.
\\ \hline
\end{tabular}
\caption{\cudr data samples for PDTB framework: counterfactual \textcolor{orange}{$Arg'_2$} being coherent with original \textcolor{blue}{$Arg_1$} and counterfactual \textcolor{orange}{$Conn'$}.}
\label{tab:cudr_samples_pdtb}
\end{table}

\begin{table}[h!]
\small
\begin{tabular}{|p{7.2cm}|}
\hline
\textbf{\cudr data samples for GDTB framework:} \\[0.5ex]

\textbf{Sample 1: } \\
\textcolor{blue}{$Arg_1$}: Due to its remarkable biodiversity, with over a third of the local plant species found nowhere else, Socotra has been designated a UNESCO World Heritage Site  \\
\textcolor{blue}{$Arg_2$}: With over 40,000 inhabitants, though, it's not just a nature reserve\\
\textcolor{blue}{Original relation}: Comparison.Concession.Arg2-as-denier (however)  \\
\textcolor{orange}{$Conn'$}: so \\
\textcolor{orange}{$Arg'_2$}: many conservation efforts are now focused on preserving its unique ecosystems. \\ \hline

\textbf{Sample 2: } \\
\textcolor{blue}{$Arg_1$}: So this place was so cool we could have spent hours in here\\
\textcolor{blue}{$Arg_2$}: The best thing that I thought about this bookstore was that they mixed in new copies of books with used copies\\
\textcolor{blue}{Original relation}: Contingency.Cause.Result (so) \\
\textcolor{orange}{$Conn'$}: but \\
\textcolor{orange}{$Arg'_2$}: the uncomfortable seating made it difficult to stay for long, despite the incredible atmosphere surrounding us. \\ \hline

\textbf{Sample 3: } \\
\textcolor{blue}{$Arg_1$}: There are flights from Sana'a via Al Mukalla\\
\textcolor{blue}{$Arg_2$}: Yemenia Airlines offers one flight per week on Thursday morning \\
\textcolor{blue}{Original relation}: Expansion.Instantiation.Arg2-as-instance (for example) \\
\textcolor{orange}{$Conn'$}: because \\
\textcolor{orange}{$Arg'_2$}: the airport reopened after extensive renovations
\\ \hline
\end{tabular}
\caption{\cudr data samples for GDTB framework: counterfactual \textcolor{orange}{$Arg'_2$} being coherent with original \textcolor{blue}{$Arg_1$} and counterfactual \textcolor{orange}{$Conn'$}.}
\label{tab:cudr_samples_gdtb}
\end{table}

\begin{table}[h!]
\small
\begin{tabular}{|p{7.2cm}|}
\hline
\textbf{\cudr data samples for RST framework:} \\[0.5ex]

\textbf{Sample 1: } \\
\textcolor{blue}{$Arg_1$}: that cultural behaviors are not genetically inherited from generation to generation \\
\textcolor{blue}{$Arg_2$}: must be passed down from older members of a society to younger members\\
\textcolor{blue}{Original relation}: adversative-antithesis (however) \\
\textcolor{orange}{$Conn'$}: specifically \\
\textcolor{orange}{$Arg'_2$}: they are learned through social interactions and environmental influences \\ \hline

\textbf{Sample 2: } \\
\textcolor{blue}{$Arg_1$}: I came up with an individual story called Thad 's World Destruction and , she wanted to illustrate it \\
\textcolor{blue}{$Arg_2$}: that 's the way we ended up doing it\\
\textcolor{blue}{Original relation}: causal-result (so)\\
\textcolor{orange}{$Conn'$}:  but\\
\textcolor{orange}{$Arg'_2$}: she thought it was too dark for children \\ \hline

\textbf{Sample 3: } \\
\textcolor{blue}{$Arg_1$}: fisherman first noticed the people\\
\textcolor{blue}{$Arg_2$}: a warship was deployed to retrieve them\\
\textcolor{blue}{Original relation}: joint-sequence (then)\\
\textcolor{orange}{$Conn'$}: because \\
\textcolor{orange}{$Arg'_2$}: he heard their laughter nearby
\\ \hline
\end{tabular}
\caption{\cudr data samples for RST framework: counterfactual \textcolor{orange}{$Arg'_2$} being coherent with original \textcolor{blue}{$Arg_1$} and counterfactual \textcolor{orange}{$Conn'$}.}
\label{tab:cudr_samples_rst}
\end{table}

\begin{table}[h!]
\small
\begin{tabular}{|p{7.2cm}|}
\hline
\textbf{\cudr data samples for SDRT framework:} \\[0.5ex]

\textbf{Sample 1: } \\
\textcolor{blue}{$Arg_1$}: the deal mechanism 's a bit clunky  \\
\textcolor{blue}{$Arg_2$}: the key is to make sure you've checked the right colour box :D\\
\textcolor{blue}{Original relation}: Contrast (by comparison)  \\
\textcolor{orange}{$Conn'$}: specifically \\
\textcolor{orange}{$Arg'_2$}: it often requires multiple steps and lengthy approvals to finalize transactions \\ \hline

\textbf{Sample 2: } \\
\textcolor{blue}{$Arg_1$}: you drive a hard bargain \\
\textcolor{blue}{$Arg_2$}: that price is too good\\
\textcolor{blue}{Original relation}: Explanation (because) \\
\textcolor{orange}{$Conn'$}:  by comparison\\
\textcolor{orange}{$Arg'_2$}: others settle for less \\ \hline

\textbf{Sample 3: } \\
\textcolor{blue}{$Arg_1$}: yep saturday 's looking promising\\
\textcolor{blue}{$Arg_2$}: saturday evening good for me too\\
\textcolor{blue}{Original relation}: Parallel (and) \\
\textcolor{orange}{$Conn'$}: because \\
\textcolor{orange}{$Arg'_2$}: the weather forecast predicts sunshine
\\ \hline
\end{tabular}
\caption{\cudr data samples for SDRT framework: counterfactual \textcolor{orange}{$Arg'_2$} being coherent with original \textcolor{blue}{$Arg_1$} and counterfactual \textcolor{orange}{$Conn'$}, while the arguments are shorter than PDTB. }
\label{tab:cudr_samples_sdrt}
\end{table}

\section{Implementation Details}

\subsection{Model fine-tuning}
\label{app: cudr model finetune}

\begin{table}[h]
\small
\centering
\setlength{\tabcolsep}{8pt} 
\renewcommand{\arraystretch}{1.2}
\begin{tabular}{ccccc}
\hline
& \multicolumn{2}{c}{\textbf{Accuracy}} & \multicolumn{2}{c}{\textbf{Logit Diff}} \\
\cline{2-5}
 & \textbf{Ori} & \textbf{CF} & \textbf{Ori} & \textbf{CF} \\
\hline
\textbf{Random Model}       & 0.50 & 0.50 & 0.00  & 0.00 \\
\textbf{Ideal Model}       & 1.00 & 1.00 & $+$  & $+$ \\
\textbf{GPT$_\text{NSP}$}   & 0.46 & 0.58 & $-$0.61 & 2.01 \\
\textbf{GPT$_\cudr$} & \textbf{0.80} & \textbf{0.80} & \textbf{7.07} & \textbf{6.59} \\
\hline
\end{tabular}
\caption{\textbf{Performance on the \cudr task:} Accuracy and logit difference are reported for each model under both original (Ori) and counterfactual (CF) scenarios.}
\label{tab:pretrain_cudr}
\end{table}

The \cudr task imposes two key requirements: \textbf{(1) Instruction following:} the model must adhere to the task format by choosing between $Arg_2$ and $Arg_2'$; and \textbf{(2) Discourse comprehension:} it must interpret the discourse relation to select the continuation that matches the given connective.
These requirements challenge the widely used GPT-2 model \cite{conmy2023towards, yao2024knowledge}. To address (1), we fine-tune GPT-2 medium on a next sentence prediction (NSP) task formatted as \cudr: selecting the correct $Arg_2$ over a mismatched $Arg_2'$ from PDTB. Without this, the model often generates irrelevant outputs.
Despite this training, GPT$_\text{NSP}$ performs poorly on the actual \cudr task, with near-random accuracy (0.46 and 0.58; see Table~\ref{tab:pretrain_cudr}).
To address (2), we further fine-tune it on strictly held-out set of PDTB data, resulting in GPT$_\text{\cudr}$, which achieves 0.80 accuracy and a significantly larger logit margin. This ensures the model is sensitive to discourse relations, making it suitable for activation patching with \cudr. These results also reflect the quality of our dataset. GPT$_\text{NSP}$ performs better on counterfactual instances than original ones (0.58 vs. 0.46 accuracy), suggesting that the counterfactual data is not only valid but also easier to interpret. The final GPT$_{\cudr}$ achieves balanced performance across both Ori and CF directions. 
Most discourse relations perform around 0.80 accuracy, with Expansion.Conjunction notably higher than 0.90. This is expected, as Expansion.Conjunction is a ``default'' continuation relation that is easier to model (also observed in pre-LLM studies). Our faithfulness metric helps normalize these raw differences by comparing activation patching outcomes to those of the (imperfect) full model, reducing the impact of absolute accuracy. 

\subsection{Computation Resources}

Our experiments on the GPT-2 medium model are conducted on a server with four NVIDIA L40 GPUs (48GB RAM each). To accelerate circuit discovery, we use the implementation by \citet{miller2024transformer} 
\footnote{\url{https://github.com/UFO-101/auto-circuit}} 
for the Edge Attribution Patching (EAP) method \cite{syed-etal-2024-attribution, nanda2023attribution}, which completes discovery for a single discourse relation in about one minute using a sample size of 32 on a single GPU. This is substantially faster than the Automatic Circuit DisCovery (ACDC) method \cite{conmy2023towards}\footnote{\url{https://github.com/ArthurConmy/Automatic-Circuit-Discovery}}.
For our indicative evaluation on GPT-2-large, experiments were run on NVIDIA A100 GPUs (80 GB RAM). For both models, we used a batch size of 1 and aggregated results over 32 samples, as activation patching is highly memory-intensive. Exploring lower-precision computation could further reduce memory demands. Ultimately, memory-efficient approaches will be crucial for scaling \cudr and circuit discovery to larger (vision-)language models such as Llama \cite{grattafiori2024llama} and Qwen \cite{bai2023qwen}.

\subsection{Scaling to Larger Models}
\label{app: scaling}

\begin{figure}[t!]
    \centering
    \includegraphics[width=\columnwidth]{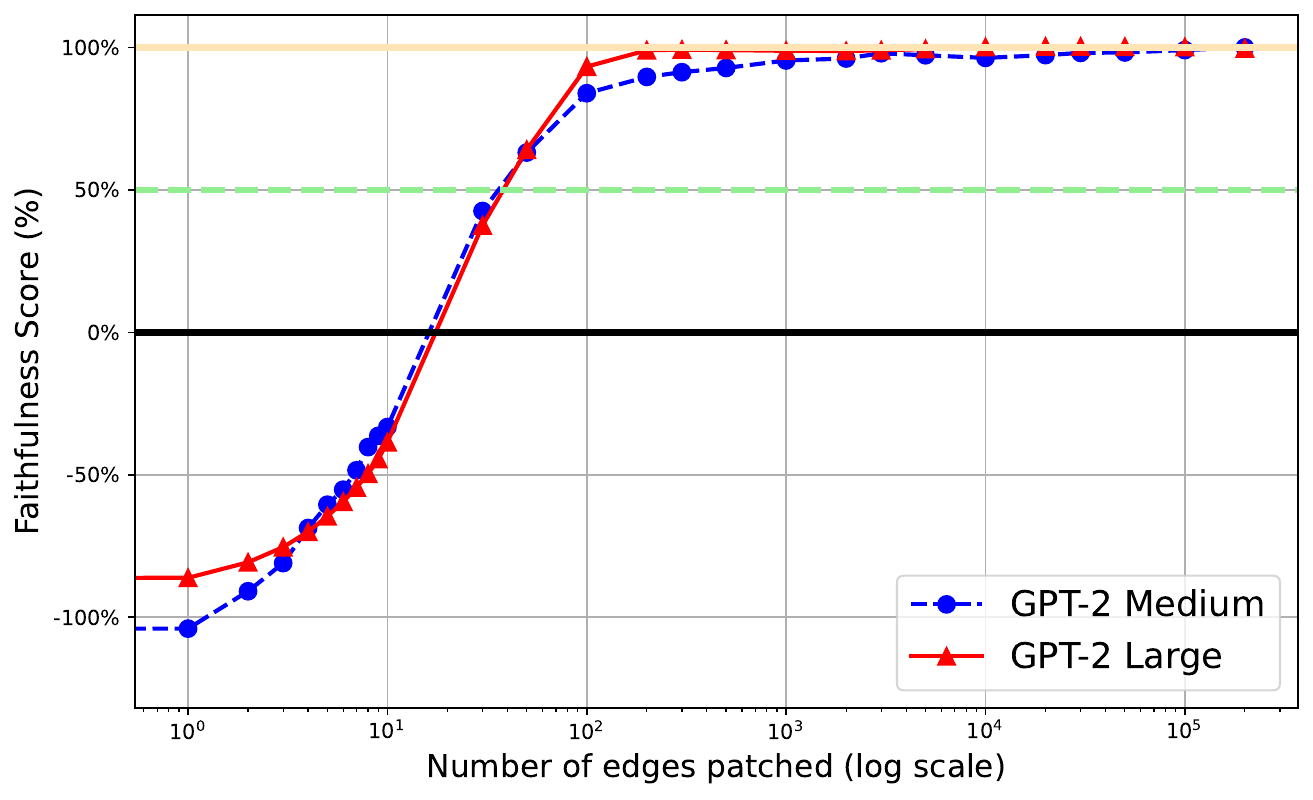}
    \caption{Comparison of GPT-2 large and GPT-2 medium models on the \cudr task.}
    \label{fig:scale-large}
\end{figure}

We replicate our experiments on the GPT-2 large model as an indicative evaluation across model sizes. Following the same recipe, we first fine-tune GPT-2 large on our \cudr task and then apply activation patching to identify circuits. Figure~\ref{fig:scale-large} shows the performance of L3-level circuits in both models (primary evaluation; other edges exhibit similar trends). Our findings indicate that (1) discursive circuits remain effective in the larger model, and (2) both models display similar trends, though GPT-2 large achieves strong performance with fewer edges. This is likely because larger models possess greater capacity for discursive understanding, with a small number of edges carrying important functions for discourse processing.

\subsection{Human-written Counterfactual} 
\label{app: human annotated cf}

\begin{figure}[t!]
    \centering
    \includegraphics[width=\columnwidth]{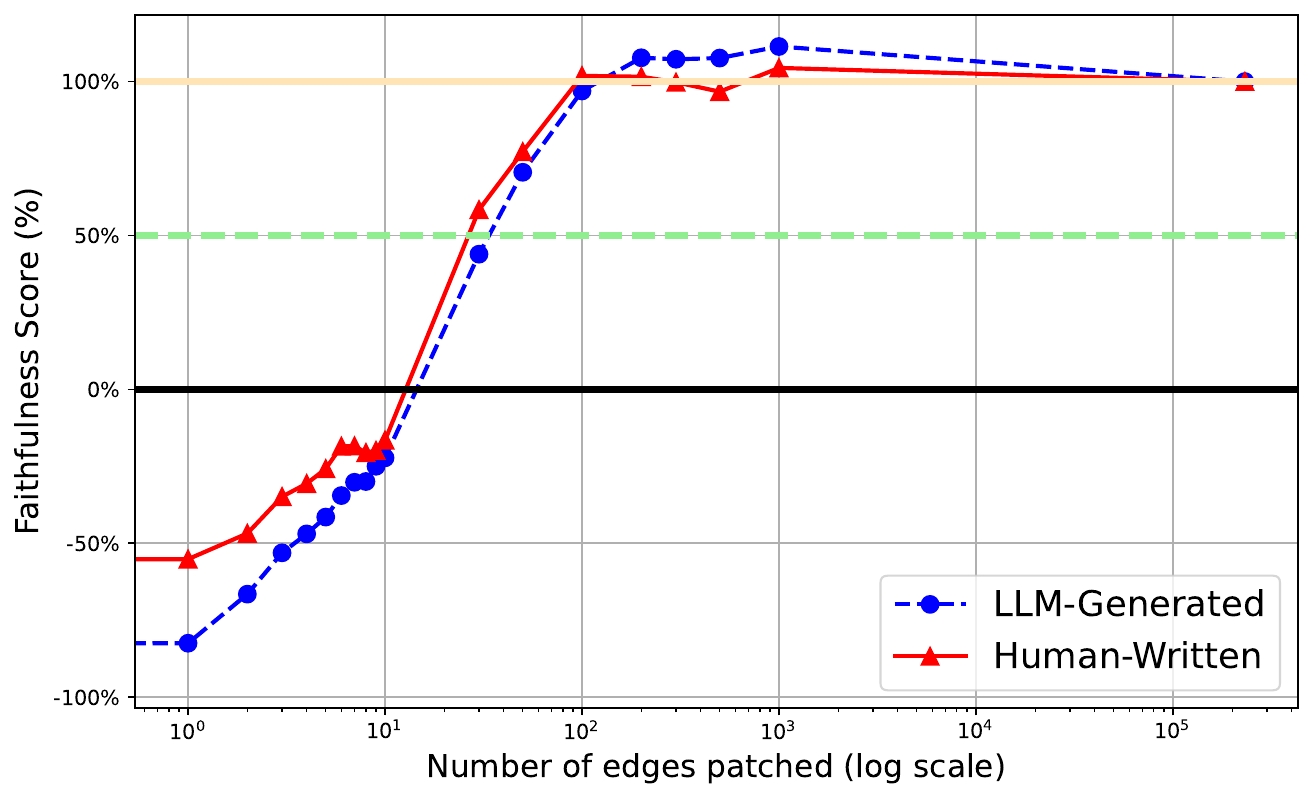}
    \caption{Comparison of circuit performance on LLM-generated and human-written \cudr data.}
    \label{fig:human-annotation-cf}
\end{figure}

The counterfactual $Arg'{2}$ instances in the \cudr datasets are generated by an LLM. To assess their quality, we additionally create a small set of human-written counterfactual $Arg'{2}$ for indicative comparison. Specifically, the first author wrote five instances for each of the 13 PDTB discourse relations, yielding a total of 65 counterfactual $Arg'{2}$.
We then evaluated model performance on the \cudr task using both LLM-generated and human-written counterfactuals. Figure~\ref{fig:human-annotation-cf} shows that the two data series follow similar trends: both initially move in the opposite direction (predicting the counterfactual $Arg'{2}$) and then, after patching around 100 edges from the clean input, recover the performance of the full model. In this indicative evaluation, the LLM-generated data aligns well with the human-written data.

\subsection{Details for Circuits Analysis Experiments}
\label{app: detail experiments}

\begin{table}[h!]
\small
\begin{tabular}{|p{7.2cm}|}
\hline

\textbf{Antonymy} \\
\textbf{Input:} The sky was \textit{bright}, far from, \textbf{Output:} dark\\
\textbf{Input:} His explanation was \textit{clear}, unlike, \textbf{Output:} confusing \\ \hline

\textbf{Coreference}\\
\textbf{Input:} \textit{John} went to the store because, \textbf{Output:} He \\
\textbf{Input:} \textit{Lisa} loves painting, and \textbf{Output:} She \\ \hline

\textbf{Negation}\\
\textbf{Input:} The answer was expected, though arrival was \textbf{Output:} delayed \\
\textbf{Input:} He expected an easy task, but it was \textbf{Output:} not \\ \hline

\textbf{Synonymy} \\
\textbf{Input:} The road was \textit{narrow}, and the alley even, \textbf{Output:} slim  \\
\textbf{Input:} The musician composed a \textit{tune}, a catchy, \textbf{Output:} melody \\ \hline

\textbf{Modality} \\
\textbf{Input:} With enough practice and support, they eventually \textbf{Output:} could  \\
\textbf{Input:} To stay healthy and fit, you \textbf{Output:} should \\

\hline
\end{tabular}
\caption{Data samples for discovering circuits for linguistic features, including antonymy, coreference, negation, synonymy, and modality. If an anchor word exists (e.g. ``John''), it was in italic form. }
\label{tab:example_linguistic_features}
\end{table}

To identify circuits responsible for linguistic features \cite{zeldes-etal-2025-erst, das2018rst}, we adopt a simplified next-word prediction setting, where the model predicts a word tied to a specific linguistic feature. This setup follows tasks like subject–verb agreement (SVA) \cite{ferrando-costa-jussa-2024-similarity} and world knowledge \cite{yao2024knowledge}. Following standard practice, we apply activation patching. The clean input is a context–target pair, while the corrupted input has the same context but a different (incorrect) target word. Activation patching identifies key edges that steer the model from the incorrect to the correct prediction. For example, for coreference, a clean input like ``Lisa loves painting'' should yield ``she''; similarly, ``John went to the store because'' should produce he'' (Table~\ref{tab:example_linguistic_features}). We patch activations from the clean input into the corrupted one to restore the correct output and identify important edges to compose the corresponding circuits. 
We select synonymy, antonymy, negation, coreference, and modality features. These features are identified as important and high-coverage discourse signals. According to the eRST taxonomy \cite{zeldes-etal-2025-erst}, synonymy (typically signaling equivalence and continuation) and antonymy (often signaling contrast) fall under the semantic category. Negation (e.g., ``not''), which frequently signals comparison relations, is also classified as a semantic signal. Coreference belongs to the reference category and is a key mechanism for maintaining discourse coherence. \citet{pitler2009automatic} find that modality features, such as modal verbs (can, should), are often associated with conditional statements that typically signal contingency relations. These categories have been shown to be prevalent across diverse genres and to support a wide range of discourse relations.

\subsection{Discursive Circuits Help Uncover Underspecified Bias}
\begin{figure}[t!]
    \centering
    \includegraphics[width=7.6cm]{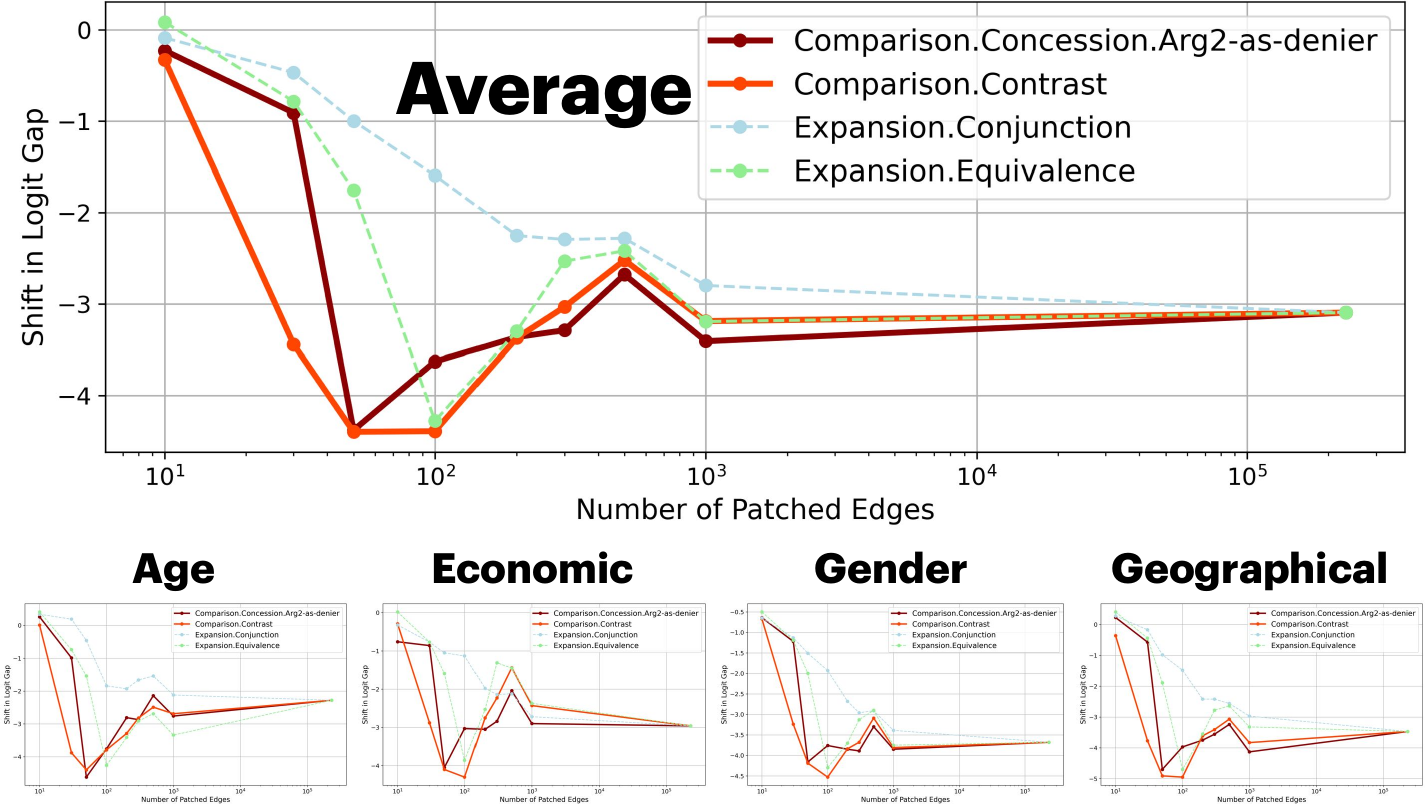}
    \caption{\textbf{Impact of discursive circuits on biased completions.} A sharper decrease in answer logit gap (Y-axis) w.r.t. patched edges (X-axis) indicates stronger circuit influence. The upper plot shows average effects. 
    }
    \label{fig:bias}
\end{figure}

The main body of the paper focuses on the \cudr task itself. To illustrate the utility of the identified discursive circuits, we present one possible use case where these circuits help reveal potential ethical biases in LLMs.
We consider scenarios where the model predicts a next sentence given an underspecified discourse relation (i.e., without an explicit connective).
For instance, ``Girls like math'' is followed by ``Boys like sports''. It is unclear whether the model interprets the two as equivalent or contrasting. Discursive circuits can uncover if the models generates the prediction for the correct reason.
To test whether the model relies on a given discursive circuit (e.g., Contrast), we destroy the activation in that circuit by patching in values from an unrelated sentence, and observe whether the output shifts toward completing that unrelated context. 
Thus, a stronger reliance results in a sharper shift.
We select four representative social biases \cite{liu2024the} and create 100 discourse instances with underspecified discourse relations. Using GPT-4o-mini, we prompt the model to generate short and simple cases that are coherent but intentionally underspecified in their discourse relation. For example, ``[A young artist painted bold lines across the canvas]$_{Arg_1}$, [A senior man updated the date in his weather journal]$_{Arg_2}$'' is a case for age bias.
Figure~\ref{fig:bias} shows output shifts under four possible biases. We find that comparison circuits produce the steepest drops (50 edges to reach bottom), indicating stronger influence. Equivalence circuits follow but require more edges (100 edges to reach bottom), while Conjunction circuits show minimal impact.
This provides mechanistic evidence that the model may exhibit a bias toward contrastive interpretations.

\subsection{Samples of Discursive Circuits Visualization}
\label{app: actual circuits} 

\begin{figure}[t!]
    \centering
    \includegraphics[width=7.6cm]{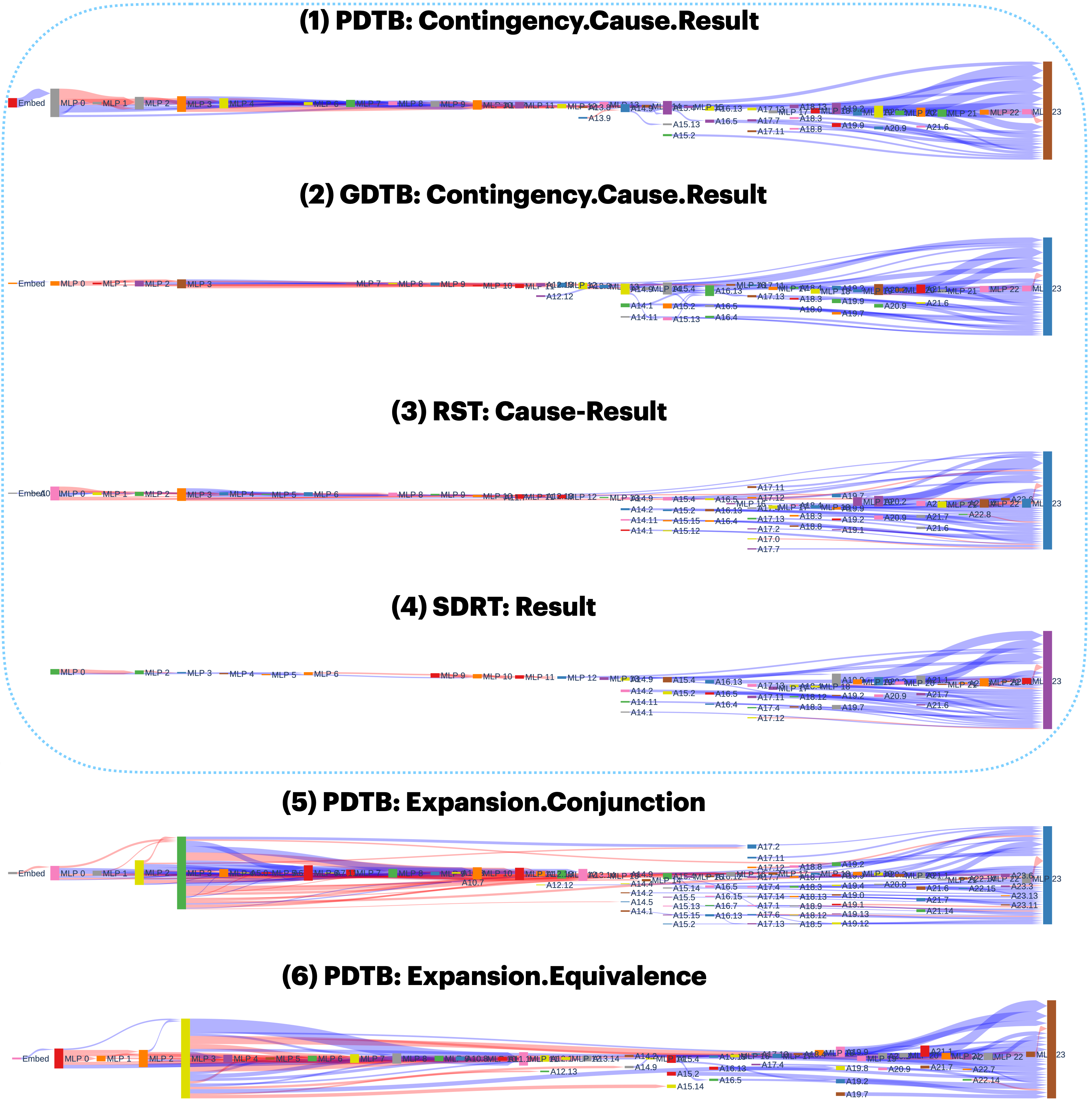}
    \caption{\textbf{Examples of discursive circuits.} Residual flows begin at the left (residual start), traverse 24 layers, and end at the right (residual end).}
    \label{fig: sample_circuits}
\end{figure}

We present representative samples of discursive circuits across different frameworks in Figure~\ref{fig: sample_circuits}. We appreciate the visualization tool created by \citet{miller2024transformer}. The left side marks the start of the residual flow from the embedding layer, continuing through 24 layers to the residual end. Each edge represents a connection between modular blocks (either MLPs or attention heads) in the transformer.
The 1st to 4th samples (highlighted by the blue dotted lines) correspond to contingency-like relations across the PDTB, GDTB, RST, and SDRT datasets. These circuits show a consistent pattern: a narrow, focused flow at the start that begins to build specialized representations from Layer 14 onward, dispersing toward the residual end. This aligns with our findings in Section~\ref{sec: linguistic features}, where discourse-specific information emerges in higher layers.
In contrast, PDTB's Expansion.Conjunction and Expansion.Equivalence (5th and 6th) are more straightforward relations (Section~\ref{sec: main exp}). Their circuits resemble an ``H'' shape, with dense processing at both the beginning and end. Together, these visualizations highlight both the commonalities and divergences in circuit structure across discourse relations.

\end{document}